\documentclass[journal]{IEEEtran}
\usepackage{float}
\usepackage{amsmath,amssymb}
\usepackage{multirow}
\usepackage{subcaption}
\usepackage{graphicx}
\begin{document}
%
\title{Remembering What Is Important: A Factorised Multi-Head Retrieval and Auxiliary Memory Stabilisation Scheme for Human Motion Prediction}
%
%
%

\author{Tharindu~Fernando,~\IEEEmembership{Member,~IEEE,} 
        Harshala~Gammulle,~\IEEEmembership{Member,~IEEE,} 
        Sridha~Sridharan,~\IEEEmembership{Life Senior Member,~IEEE,}
        Simon~Denman,~\IEEEmembership{Member,~IEEE,}
         ~and~ Clinton~Fookes,~\IEEEmembership{Senior Member,~IEEE.}

\IEEEcompsocitemizethanks{\IEEEcompsocthanksitem T. Fernando, H. Gammulle, S. Sridharan, S.Denman, and C. Fookes are with The Signal Processing, Artificial Intelligence and Vision Technologies (SAIVT), Queensland University of Technology, Australia.\protect }}

\markboth{}%
{Fernando \MakeLowercase{\textit{et al.}}: A Factorised Multi-Head Retrieval and Auxiliary Memory Stabilisation Scheme for Human Motion Prediction}

\maketitle

\begin{abstract}
Human's exhibit complex motions that vary depending on the task that they are performing, the interactions they engage in, as well as subject-specific preferences. Therefore, forecasting future poses based on the history of the previous motions is a challenging task. This paper presents an innovative auxiliary-memory-powered deep neural network framework for the improved modelling of historical knowledge. Specifically, we disentangle subject-specific, task-specific, and other auxiliary information from the observed pose sequences and utilise these factorised features to query the memory. A novel Multi-Head knowledge retrieval scheme leverages these factorised feature embeddings to perform multiple querying operations over the historical observations captured within the auxiliary memory. Moreover, our proposed dynamic masking strategy makes this feature disentanglement process dynamic. Two novel loss functions are introduced to encourage diversity within the auxiliary memory while ensuring the stability of the memory contents, such that it can locate and store salient information that can aid the long-term prediction of future motion, irrespective of data imbalances or the diversity of the input data distribution. With extensive experiments conducted on two public benchmarks, Human3.6M and CMU-Mocap, we demonstrate that these design choices collectively allow the proposed approach to outperform the current state-of-the-art methods by significant margins: $>$ 17\% on the Human3.6M dataset and $>$ 9\% on the CMU-Mocap dataset. 
\end{abstract}


\IEEEpeerreviewmaketitle

\section{Introduction}

\IEEEPARstart{I}{n} real-world day-to-day activities human exhibit complex and highly varied poses. For instance, considering the motion as a person walks, depending on objects that the person is carrying, and any interactions that take place, we may observe diverse variations between different human motion sequences \cite{ionescu2013human3}. Moreover, subject specific details such as limb lengths, and skeleton structure result in nuances which hamper attempts to predict future skeleton motion \cite{ionescu2013human3}. As such, there are global covariance factors across different scenes, camera setups, and tasks that the predictive algorithms should compensate for when predicting future motion. In addition, subject-specific local variations within the same task, scene, and camera setups should also be considered for accurate forecasting. 

\begin{figure}
    \centering
    \includegraphics[width=\linewidth]{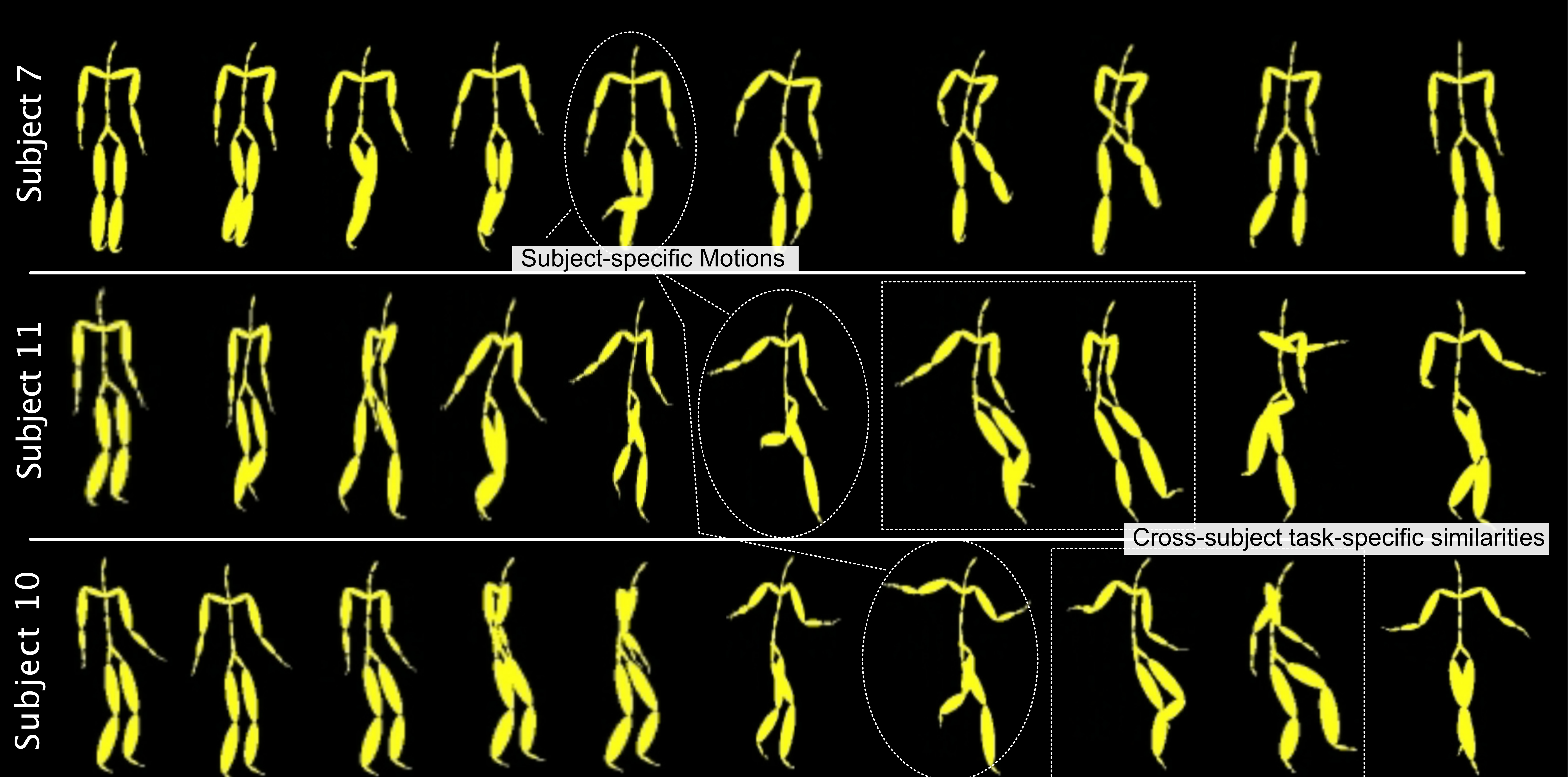}
    \caption{Visual illustration of subject specificity and cross-subject task-specific similarities using samples for soccer action taken from the CMU-Mocap dataset. The subject specificity is illustrated with ellipses while the cross-subject task-specific similarities are shown with squares.}
    \label{fig:figure_1}
\end{figure}

Existing state-of-the-art algorithms have adapted multi-scale joint pooling or multi-scale skeleton segmentation as a method to compensate for global and local influencing factors \cite{mao2019learning, li2020dynamic, dang2021msr}. We acknowledge the fact that such multi-scale modelling would help model distinct structural dependencies within human motion patterns. However, such a modelling approach fails to explicitly capture subject-specific, task-specific, and global factors that influence distinct motion patterns. Explicitly capturing these factors will help to capture semantically meaningful features, yielding better forecasting accuracy. Fig. \ref{fig:figure_1} illustrates some examples from the CMU-Mocap dataset \footnote{http://mocap.cs.cmu.edu/} that show subject specificity and cross-subject task-specific similarities that the learning framework can leverage; yet the factorisation of subject-specific, task-specific and other influential factors using the existing human motion modelling framework is an intricate task. In addition, the prevailing feed-forward deep learning pipeline that is leveraged within the state-of-the-art motion modelling literature does not allow for comparisons across samples that the model has previously seen during training, which is vital for the effective separation of salient task-specific and subject-specific attributes. 

This paper proposes a novel deep neural network framework which effectively disentangles subject-specific, task-specific and other auxiliary features from human motion representations. In addition, we design an innovative auxiliary memory-powered multi-head retrieval scheme to efficiently incorporate these factorised details for knowledge retrieval. Specifically, our auxiliary memory allows us to perform querying of the features from historically observed motion sequences in relation to different factorised features. For example, we can query the history in relation to the task that the subject of interest is performing, the observations that our model has seen which were performed by the same subject, or the observations related to the same background auxiliary features, and efficiently combine them for our predictive task at hand. Moreover, we make this feature disentanglement process dynamic where the feature masks that we used to factorise the features can dynamically change depending on the input, even after model training. 

Several prior works have demonstrated the need for specialised losses to optimise the memory update procedure to work in situations with highly imbalanced data distributions. Relying on the auxiliary memory update procedure to be optimised from the classification or regression objective alone could lead the auxiliary memory to memorise only the information from the frequent classes. To alleviate this issue we propose a novel loss function which rewards diverse auxiliary memory content and penalises similar content. Moreover, smaller auxiliary memory sizes and/or high diversity in the input data distribution can lead to frequent updates and unstable auxiliary memory content. As such, we additionally propose an innovative loss function which penalises large memory updates that occur after the content of a particular auxiliary memory slot has been stabilised or consolidated, which we measure based on the time that content is initially written to the auxiliary memory and the frequency of the updates of that particular memory slot. 

To the best of our knowledge, this is the first work to combine feature factorisation and auxiliary memory-powered knowledge retrieval. The main technical contributions of this paper, through which we introduce the proposed Factorised Multi-Head retrieval and Stabilisation based Auxiliary Memory (FMS-AM), can be summarised as follows. 

\begin{enumerate}
    \item We introduce a novel feature factorisation strategy to disentangle subject-specific, task-specific, and auxiliary attributes from human motion representations.
    \item We present a new mask generation strategy that induces dynamicity in mask generation.
    \item We propose an innovative multi-head access-based feature querying strategy to effectively retrieve knowledge embedded within an auxiliary memory.
    \item Two novel loss functions are proposed to encourage diversity and stabilisation of the memory content while encouraging continual learning of important facts. 
\end{enumerate}

\section{Related Work}
In this section, we summarise the related works of this article which we categorise into works on Skeleton-based Human Motion Prediction (Sec. \ref{sec:litreature_review_motion_prediction}), and literature on Auxiliary Memory Powered Neural Networks (Sec. \ref{sec:litreature_auxiliary_memory}).

\subsection{Skeleton-based Human Motion Prediction}\label{sec:litreature_review_motion_prediction}
3D Skeleton-based human motion prediction is considered a fundamental research topic that benefits a myriad of application areas including intelligent surveillance, autonomous driving, and human-robot interaction. A variety of human motion prediction methods have been proposed that range from traditional machine learning-based methods \cite{taylor2009factored, taylor2006modeling, wang2005gaussian} to deep learning-based methods \cite{fragkiadaki2015recurrent, gui2018adversarial, guo2019human, li2020dynamic, zhou2021learning, dang2021msr}. Most recent success within this domain has been achieved using Graph Neural Networks (GNNs), with which researchers have exploited the relationships and constraints between different body components. Specifically, the Dynamic Multiscale Graph Neural Networks (DMGNN) \cite{li2020dynamic} architecture models the human body in a multi-scale graph in which nodes are body components at various scales, and edges represent pairwise relations between those components. The multi-Scale Graph Computational Unit (MGCU) is the main processing component within the DMGNN architecture in which single-scale graphs extract features at their respective scales, and these extracted features are passed through different scale graphs to connect body components across two scales. Graph connections are initialized using predefined physical connections and are adaptively adjusted during model training. Then a graph-based GRU (G-GRU) is employed to predict future poses using the extracted representations. The MGCN \cite{zhou2021learning} architecture follows a similar structure to DMGNN where the authors have utilised a Scale Interactional Module (SIM) to encode the human pose at multiple scales. Similarly, the MSR-GCN \cite{dang2021msr} architecture is motivated by the concept of coarse to fine-grained prediction generation. The descending GCN blocks are used to abstract the human pose at four levels of 12, 7, and 4 joints respectively, after which ascending GCN blocks reconstruct future poses at increasingly fine-grained scales. More recently, a multi-stage prediction framework named Spatial Dense Graph Convolutional Networks (S-DGCN) is proposed in \cite{ma2022progressively}. In this progressively evolving architecture, the observed pose sequence and the initial guess of the future pose, which is predicted by the previous stage, are taken as the input. The authors show that this recursively updating approach leads to the progressive improvement of the initial guess predicted by the first stage. 

In a different line of work, \cite{zhong2023geometric} proposed a Geometric Algebra-based Multi-view Interaction network (GA-MIN) leveraging Geometric Algebra tools to mitigate feature similarity issues when aggregating high-dimensional features using a deep and multi-stage GCN. Specifically, the authors use a Graph Spectrum Self-interaction module to discover the repeated motion within the input sequences, and use a Graph Spectrum Global-interaction module to extract informative motion representations within spectrum bands. The architecture of PK-GCN \cite{sun2022overlooked} is motivated by the observation that sequence interpolation is easier than extrapolation. As such two networks, the named InTerPolation learning Network (ITP-Network) and Final Prediction Network (FP-Network), are proposed. In contrast to an approach where we directly extrapolate the relationship between the observed sequence and the target, the ITP-Network learns to encode the input and a privileged sequence to interpolate the in-between frames in the predicted sequence. The FP-Network receives the encoded input but the privileged sequence is not visible to it. It uses a PK-Simulator that distills the privileged sequence based on the observed sequence. As such the FP-Network is able to imitate the interpolation process. The Dynamic Pattern-based collaborative modeling network (DPnet) \cite{tang2023collaborative} considers preserving dynamic information of the joints. The authors show that global modeling of joint relationships could lead to the introduction of undesired trajectory constraints. To address this issue the authors propose a keyframe-enhanced module that augments the extracted temporal features by encoding the input into different-length sub-sequences. The dynamic patterns of different joints are discriminated using a dynamic pattern-guided feature extractor. 

Considering the recent success of Transformer Networks in numerous sequence modeling tasks \cite{li2022contextual, fan2022point, vaswani2017attention}, they also have been used to model human motion sequences \cite{aksan2021spatio, cai2020learning}. The self-attention mechanism within the transformer has been adapted to compute the pair-wise joint relationships. However, the authors of \cite{ma2022progressively} have shown that GCNs are more robust and efficient compared to transformers in modeling the pairwise relations of joints. As such, we adopt a GCN-based backbone to encode input motion sequences. However, in contrast to existing works in skeleton-based human motion prediction, we leverage a novel feature factorisation strategy to disentangle subject-specific, task-specific, and other auxiliary features from the encoded inputs. Furthermore, a novel auxiliary memory architecture is proposed to query subject-specific historical local patterns, as well as global task-specific patterns, that are embedded within the memory architecture. 


\subsection{Auxiliary Memory Powered Neural Networks}\label{sec:litreature_auxiliary_memory}
Auxiliary Memory-Powered Neural Networks (AMNNs) \cite{priyasad2022detecting, priyasad2021memory, fernando2020detection, fernando2020neural, yang2019visual, nguyen2023memory, park2018towards, park2018quantized} are a recent and pivotal development within deep learning. They have shown tremendous success in automatically deriving long-term dependencies between input observations, and have been able to attain state-of-the-art results in various machine learning tasks, including, anomaly detection \cite{fernando2020detection, priyasad2022detecting, fernando2020neural}, interaction modelling \cite{nguyen2023memory}, multimodal data fusion \cite{priyasad2021memory, park2018towards}, visual tracking \cite{yang2019visual}, visual question answering \cite{khademi2020multimodal}, and human action recognition \cite{gammulle2019forecasting}. 

Specifically, an AMNN utilises explicit storage (memory) to store important facts and automatically retrieve relevant long-term dependencies when making a decision regarding a particular input, which is highly beneficial when extrapolating into the distant future. AMNNs fall under the category of stateful neural networks which maintain and temporarily evolve their states across the entire training/testing phase, in contrast to typical feed-forward and recurrent neural network architectures which only map the relationships within a particular input. This statefulness offers a greater utility for modeling relationships across different data elements in the dataset, enabling elevated levels of knowledge extraction. 

There are three primary functions that facilitate this temporal evolution of the knowledge captured in the AMNN. A query function (also called an input controller), which is composed of trainable neural network layers, transforms the input embedding into a vector to query the memory. Using this query vector, the similarity between the content of each memory slot and the query is measured and relevant memory slots for knowledge retrieval are identified. A composer function is used to transform the content retrieved from the identified memory slots into the memory output. The final task is to update the auxiliary memory content and propagate it into its next state. A memory update/write function receives the current memory output and it generates a vector to update the memory. Then the content of the slots which we leveraged in the memory read are updated using the generated memory update vector. It can be seen that the query function plays a pivotal role in identifying salient content in the memory which relates to the current input and can aid the task at hand. However, we observe that the current single-head access schemes used by such methods limit the identification of salient memory slots, as it focuses on the entire content of input and a particular memory slot. But in applications such as human motion synthesis, there are numerous task-specific and subject-specific factors that could be represented to various degrees within the embeddings. As such, a multi-head retrieval scheme is preferred in which varying levels of attention can be paid to these numerous influential factors. To the best of our knowledge, this is the first work to propose such a multi-head knowledge retrieval scheme in AMNNs. Furthermore, we propose two novel objective functions which encourage diversity in the memory content while also encouraging the memory content to be stable (discouraging frequent updates).

\section{Methods}

In this section we outline our proposed approach. We first introduce the encoder that we use to encode the input pose sequence (see Sec. \ref{sec:encoder}). Sec. \ref{sec:factorisation} discuss the feature factorisation strategy that we implement to disentangle the encoded features. In Sec. \ref{sec:mask_generation} we present our pipeline for generating dynamic masks based on the encoded information in the input feature, and sections \ref{sec:auxiliary_memory} and \ref{sec:memory_update} present the proposed multi-head knowledge retrieval and auxiliary memory stabilisation procedures respectively, within the proposed auxiliary memory module. The pose sequence prediction process using our decoder is presented in Sec. \ref{sec:decoder}. Finally, the implementation details of the framework are discussed in Sec. \ref{sec:implementation}.

\subsection{Multi-scale GCN Encoder}\label{sec:encoder}
Input to our encoder is a sequence of human poses, and our encoder transforms this to a deep representation by hierarchically encoding it at multiple scales. In this subsection, we explain this hierarchical encoding process.

Formally, let $X_{1:T_{obs}} = [x_1, x_2, \ldots, x_{T_{obs}}]$ consist of $T_{obs}$ consecutive human pose observations, where $x_i \in \mathbb{R}^{K}$ and $K$ is the data dimension that we utilise to describe each pose observation. $x_i \in \mathbb{R}^{K} \in \mathbb{R}^{J \times D}$ represents a single human pose which is composed of $J$ joints, and each joint is represented in $D$-dimensional space. In the datasets used in this work pose is observed in 3 dimensions, thus $D=3$. Our objective is to anticipate future poses for the duration $T_{obs}$ + 1 to $T_{obs} + T$. We denote the predicted pose sequence as $\hat{Y}_{T_{obs + 1}: T_{obs} + T} = [\hat{y}_1, \hat{y}_2, \ldots, \hat{y}_T]$, and the ground truth observation as ${Y}_{T_{obs + 1}: T_{obs} + T} = [{y}_1, {y}_2, \ldots, {y}_T]$.

Inspired by the recent success of graph neural networks to model the spatial structure of the poses \cite{dang2021msr, li2020dynamic} and the proven ability of the graph convolution operation to retrieve spatial and structural dependencies between human joints, we utilise a Graph Convolution Network (GCN) as our backbone to extract features from the input pose sequence. Following \cite{dang2021msr, li2020dynamic} we first replicate the last pose observation, $x_{T_{obs}}$, $T$ times making the input sequence of length $T_{obs} + T$. Similar to \cite{dang2021msr, li2020dynamic} we represent pose as a fully connected graph with $K$ nodes and the adjacency matrix, $A \in \mathbb{R}^{K \times K}$, which is learned during the training process represents the strength of the dependencies between pairs of joints.

Our GCN-based pose encoder is composed of $L$ graph stacked convolution layers. At each level, $l \in [1, 2, \ldots, L]$, the output, $H^{l+1}$, can be defined as,
\begin{equation}
    H^{l+1} = f_{GCN} (A^l, H^l, W^l),
\end{equation}
where $F_{GCN}$ is the activation function and $W^l \in \mathbb{R}^{F^l \times F^{l+1}}$ are the trainable parameters of the graph convolution layer. $H^{l+1} \in \mathbb{R}^{K \times F^{l+1}}$ is the output which is passed to the next graph convolution layer. $F^l$ denotes the embedding dimension of the layer $l$. 

Motivated by the success of \cite{dang2021msr, li2020dynamic} in leveraging multi-scale GCNs for capturing the hierarchical spatial and structural relationships of human pose using multi-scale representations, we also employ a series of GCNs to abstract the human pose. Our descending GCN blocks downsample the resolution of the pose sequence by pooling adjacent joints. For instance, if the input pose is represented with 22 joints (i.e. $J=22$) in three dimensions (i.e. $D=3$), then the input feature space to our first downsampling block, $DN_0$ is of shape $K_0 \times F$ with $K_0 = 22 \times 3 = 66$. This input is down-sampled and the input to the second downsampling block, $DN_1$, is of shape $K_1 = 12 \times 3 = 36$. Similarly, the $\mathrm{3^{rd}}$ and $\mathrm{4^{th}}$ downsampling blocks, $DN_2, DN_3$, set $K_2= 7 \times 3 = 21$ and $K_3 = 4 \times 3 = 12$, respectively. Note that for all the downsampling blocks we use the same embedding dimension $F$. We note that this encoding process is identical to the encoder of \cite{dang2021msr}.

\subsection{Factorised Embeddings}\label{sec:factorisation}

This subsection illustrates how we disentangle the embeddings generated by the encoder introduced in Sec. \ref{sec:encoder}. Specifically, we leverage masking operations to disentangle subject-specific, task-specific, and other auxiliary features from the output of our encoder. Formally, let $z_t^{DN_3} \in \mathbb{R}^{K_3 \times F}$ denote the output of the final downsampling block, $DN_3$, for the observed pose input at time instance $t$. Then,
\begin{equation}
    \begin{split}
    z_t^{sub} & = z_t^{DN_3} \bigotimes m_t^{sub},\\
    z_t^{task} & = z_t^{DN_3} \bigotimes m_t^{task},\\
    z_t^{aux} & = z_t^{DN_3} \bigotimes m_t^{aux},
    \end{split}
\end{equation}
can be used to split the respective features across subject, task, and auxiliary segments. Here $\bigotimes$ denotes element-wise multiplication by each mask, $m_i \in \mathbb{R}^{K_3 \times F}$. It should be noted that the masking operation only occurs in the embedding dimension, $F$, and we retain all the elements in dimension $K_3$. Details regarding the mask generation process are presented in Sec. \ref{sec:mask_generation}.

To ensure that the features are properly disentangled, and the segregated embeddings capture the intended attributes for the specified segments (i.e. subject, task, etc.) and only for that specified segment, we embed additional classification objectives alongside the primary future pose sequence prediction task.

Specifically, each of the supplementary classification heads first concatenates the relevant embeddings for the entire observed sequence such that,
\begin{equation}
    \begin{split}
    Z^{sub} & = [z^{sub}_1 \bigoplus z^{sub}_2 \bigoplus \ldots \bigoplus z^{sub}_t], \\
    Z^{task} & = [z^{task}_1 \bigoplus z^{task}_2 \bigoplus \ldots \bigoplus z^{task}_t],
    \end{split}
\end{equation}
where $\bigoplus$ denotes column-wise concatenation. The resultant feature vectors, $Z^{sub}$ and $Z^{task}$, are passed through a global average pooling operation, $f_{GAP}$, which transforms the input feature vectors to $\dot{z}^{sub}$ and $\dot{z}^{task}$ respectively, where $\dot{z}^{sub} \in \mathbb{R}^{K_3 \times F}$ and $\dot{z}^{task} \in \mathbb{R}^{K_3 \times F}$. Then we pass the resultant embeddings, $\dot{z}^{sub}$ and $\dot{z}^{task}$, through the respective classifiers such that,
\begin{equation}
    \begin{split}
    \hat{y}^{sub} & = f_{SUB}(\dot{z}^{sub}), \\
    \hat{y}^{task} & = f_{TASK}(\dot{z}^{task}),
    \end{split}
\end{equation}
where $f_{SUB}$ and $f_{TASK}$ denote subject and task classification sub-networks, respectively. Note that the evaluation protocol for skeleton-based motion synthesis in most of the popular benchmarks is leave-one-subject-out. Therefore, direct classification of the subject identity is not appropriate as the model is observing completely unseen subjects during the testing phase. To alleviate this issue we construct the subject classification as a contrastive learning task where two arbitrarily sampled pose sequences, which could be of the same subject or different subjects, are presented to the model and using the two $\dot{z}^{sub}$ embeddings extracted for the two inputs the $f_{SUB}$ network classifies whether they belong to the same subject or not. We use contrastive loss \cite{yang2022robust} to optimise this task. This contrastive learning helps the model to identify salient characteristics embedded within the input pose sequences that are subject-specific which will, in turn, help the motion prediction. Furthermore, we can use this same pipeline for both the training and testing phases. For task classification, noting that we have the same set of tasks in the training and testing sets, we use categorical cross-entropy loss. 

\subsection{Dynamic Mask Generation}\label{sec:mask_generation}
\begin{figure}
    \centering
    \includegraphics[width=\linewidth]{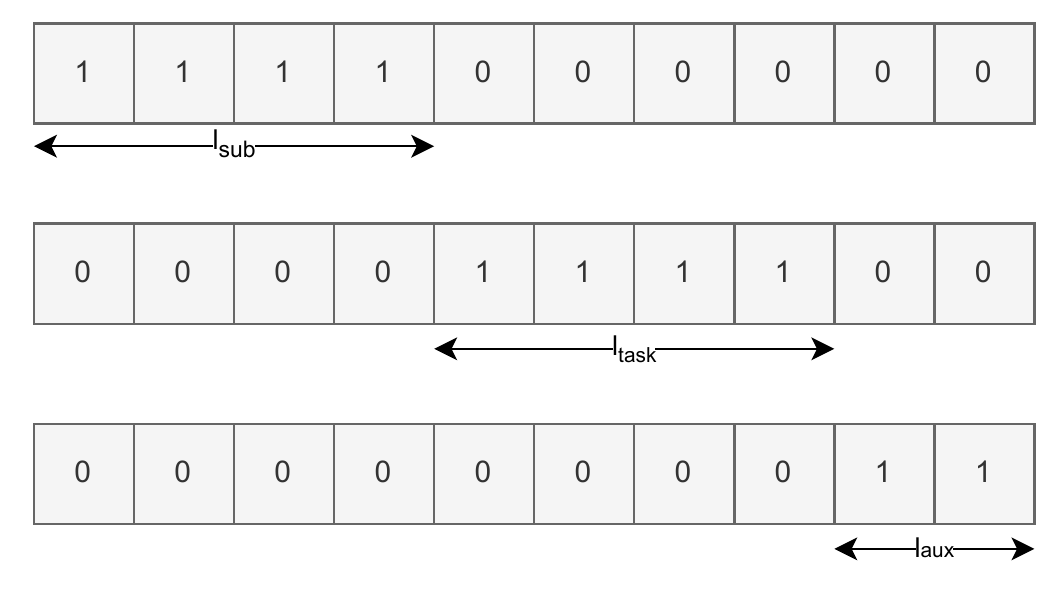}
    \caption{Illustration of fixed masks where we pre-define the regions for subject-specific, task-specific, and auxiliary features.}
    \label{fig:fixed_masks}
\end{figure}

The next task is to generate the masks such that subject, task, and auxiliary features can be disentangled. A naive way to generate the masks is to use a fixed mask such that a certain fixed region (of length $l_{sub}$) is completely dedicated to a specific subset of features, and the rest of the elements in $z_t^{DN_3}$ are completely masked out. For instance, if we select the first $l_{sub}$ elements for subject-specific features, the next $l_{task}$ for task-specific features, and the rest of the elements to carry auxiliary information, then the 3 masks can be visualised as in Fig. \ref{fig:fixed_masks}.

We see several drawbacks to this approach. First, it assumes that there exist hard/rigid boundaries between subject, task, and auxiliary segments in the feature vector and those elements are either fully active or not; i.e. it does not allow partial activation of the mask. Second, it makes the feature disentanglement process static after the model training process, and doesn't allow the feature selection to dynamically change based on the embedded information. To resolve these drawbacks we propose to dynamically generate masks based on the encoded information in $z_t^{DN_3}$. 

\begin{figure}[htbp]
    \centering
    \includegraphics[width=\linewidth]{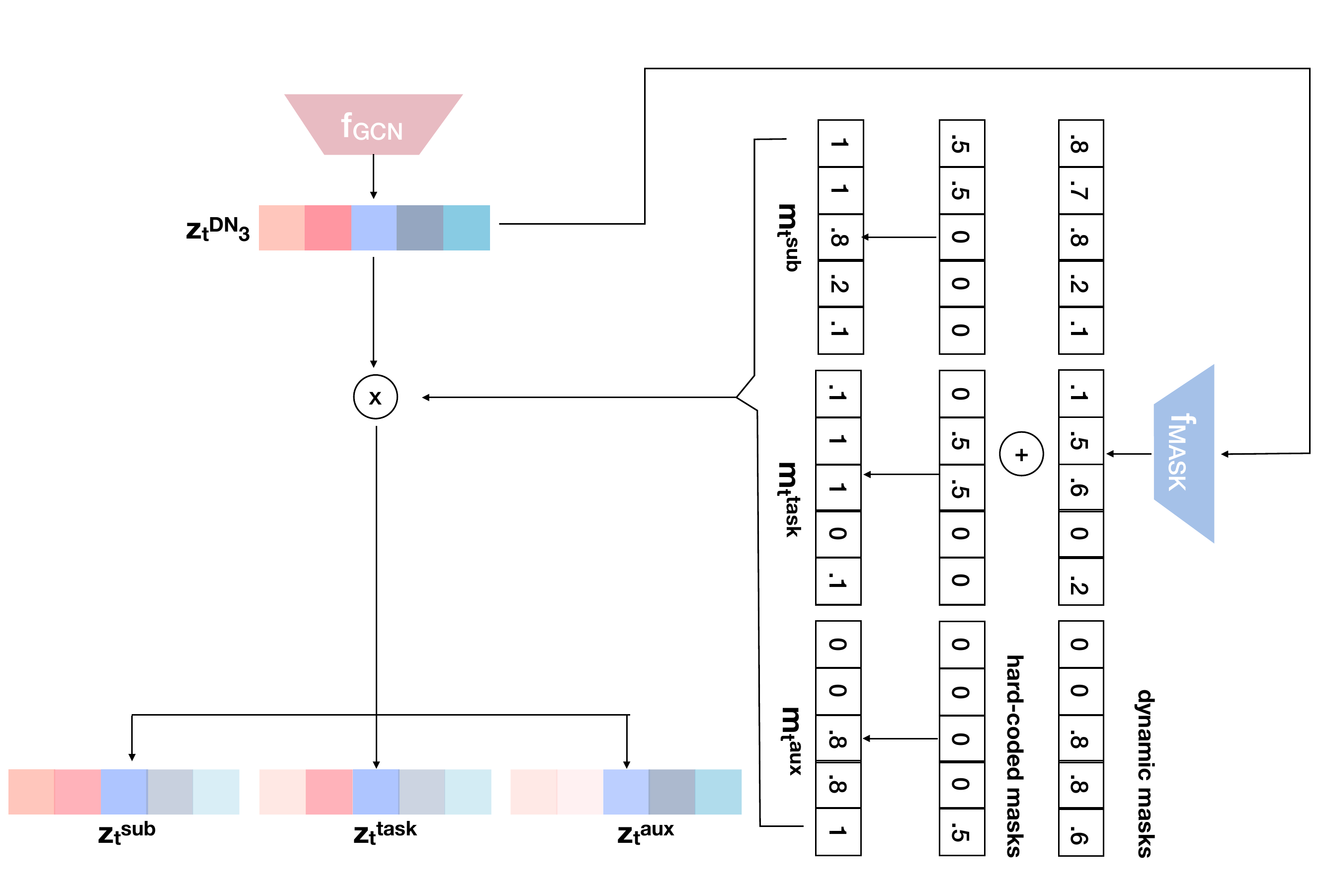}
    \caption{Dynamic Mask Generation and Feature Factorisation: Using the feature vector, $z_t^{DN_3}$, a neural network generates residuals to augment the fixed mask. These masks are then utilised to factorize $z_t^{DN_3}$.}
    \label{fig:dynamic_masking}
\end{figure}

Specifically, we utilise a set of hard-coded masks (i.e. $\dot{m}^{sub}, \dot{m}^{task}$ and $\dot{m}^{aux}$) based on the proportion in the vector $z_t^{DN_3}$ that we desire the subject, task and auxiliary features to occupy. Then, using a neural network that is parameterised by the function $f_{MASK}$ we generate residuals to augment the fixed masks,$\dot{m}^{sub}, \dot{m}^{task}$ and $\dot{m}^{aux}$. Fig. \ref{fig:dynamic_masking} visually illustrates this mask generation process.

Formally, $f_{MASK}$ outputs 3 mask vectors such that,
\begin{equation}
    (\Tilde{m}^{sub}_t, \Tilde{m}^{task}_t, \Tilde{m}^{aux}_t) = f_{MASK}(z_t^{DN_3}),
\end{equation}
where $\Tilde{m}^{sub}_t, \Tilde{m}^{task}_t$ and $\Tilde{m}^{aux}_t$ each  $\in \mathbb{R}^{K_3 \times F}$ and represent the residual masks for the subject, task and auxiliary regions respectively. Then, the augmented mask can be generated using
\begin{equation}
    \begin{split}
        m^{sub}_t & = \dot{m}^{sub} + \Tilde{m}^{sub}_t, \\
        m^{task}_t & = \dot{m}^{task} + \Tilde{m}^{task}_t, \\
        m^{aux}_t & = \dot{m}^{aux} + \Tilde{m}^{aux}_t . 
    \end{split}
\end{equation}

This can be seen as injecting plasticity \cite{fernando2020neural, paredes2019unsupervised} into a fixed set of hard-coded masks. Different initialisation functions can be used to initialise the fixed masks and for simplicity we used a uniform function such that all the elements within the particular segment (i.e. $l_{sub}, l_{task}, l_{aux}$) are initialised to a value of $0.5$ and $l_{sub} = l_{task} = l_{aux} = F/3$. Furthermore, the generated masks are normalised using the Gumbell softmax operation which is a differentiable relaxed one-hot vector-like operation. The temperature parameter, $\tau$, within the Gumbell softmax controls the sparsity of the resultant mask, with lower temperature values increasing output sparsity. We experimentally choose the value for $\tau$ which allows us to balance the fixed and plastic potions of the generated masks. 

We compare this mask generation process with the splitting network proposed in our prior work \cite{fernando2022split}. In \cite{fernando2022split} we generated the entire mask while in the proposed work we generate only the residuals for the fixed mask. We believe predicting the entire mask using Gumbell softmax operation is too restrictive as it can lead to most values in the predicted mask being zero. We experimentally compare the proposed architecture and the masking process of \cite{fernando2022split} where we demonstrate this drawback. We refer readers to the ablation evaluations in Sec. \ref{Sec:ablations} details. Furthermore, we note that to the best of our knowledge, this is the first work to embed plastic/dynamic mask generation methods within stateful, continual learning neural networks.

\subsection{Auxiliary Memory and Multi-Head Retrieval}\label{sec:auxiliary_memory}

The main goal of our feature factorisation process in Sec. \ref{sec:factorisation} is to effectively utilise this to augment the knowledge retrieval process in the auxiliary memory. As illustrated in Sec. \ref{sec:litreature_auxiliary_memory}, a typical memory architecture is composed of a memory stack with $s$ slots, and $M_{\lambda-1}$ denotes the state of the memory at time $\lambda-1$. Note that this time is measured with respect to the execution of the memory module (i.e. training/testing iteration) and not with respect to the time in input sequences (i.e. $t$). 

In a traditional memory retrieval operation we first pass the entire input to the memory, $z_t^{DN_3}$, through a query function, $f_{QUERY}$, to generate a query vector, $q^{DN_3}_{\lambda}$, to query the memory such that
\begin{equation}
    q^{DN_3}_{\lambda} = f_{QUERY}(z_t^{DN_3}).
\end{equation}
Then we retrieve memory slots that contain information related to our query using
\begin{equation}
    \beta^{DN_3}_{\lambda} = \mathrm{softmax}([q_t^{DN_3}]^\top M_{\lambda-1}),
\end{equation}
and the memory output, $\mu_{\lambda}$, at time instance $\lambda$ is computed using,
\begin{equation}
    \mu_{\lambda} = [\beta^{DN_3}_{\lambda}]^\top M_{\lambda-1}).
\end{equation}

However, we observe several limitations of the direct application of a single-head access scheme for memory knowledge retrieval in motion synthesis. In particular, when there exist numerous task-specific and subject-specific variations embedded in the same feature vector this could negatively impact the softmax score-based identification of relevant memory slots. For instance, if the memory contains information related to the same subject in the query but the tasks are different, there is a possibility that such memory slots will not be identified as relevant to the current query due to the mismatch of the task-specific features. Similarly, information for similar tasks but with different subjects may be ignored as the query operation is considering the entire query vector. This issue can be overcome using the proposed feature factorisation strategy where we can generate multiple queries using the individual factorised attributes. Specifically, we use $z^{sub}_t$ and $z^{task}_t$ in addition to $Z^{DN_3}_t$ as inputs to the memory module, and generate two additional query vectors such that,
\begin{equation}
\begin{split}
    q^{sub}_{\lambda} & = f_{QUERY, SUB}(z_t^{sub}), \\
    q^{task}_{\lambda} & = f_{QUERY, TASK}(z_t^{task}),
\end{split}
\end{equation}
which are leveraged to retrieve memory slots using
\begin{equation}
\begin{split}
    \beta^{sub}_{\lambda} & = \mathrm{softmax}([q_t^{sub}]^\top [ M_{\lambda-1}] [ m_t^{sub} \odot e^s]), \\
    \beta^{task}_{\lambda} & = \mathrm{softmax}([q_t^{task}]^\top [ M_{\lambda-1}] [ m_t^{task} \odot e^s]),
\end{split}
\end{equation}
where $e^s$ is a matrix of ones, and $\odot$ denotes the outer product which duplicates its left vector $s$ times to form a matrix. Then the knowledge retrieved from memory can be defined by, 
\begin{equation}
\begin{split}
    \mu^{sub}_{\lambda} & = [\beta_t^{sub}]^\top [ M_{\lambda-1}] [ m_t^{sub} \odot e^s], \\
    \mu^{task}_{\lambda} & = [\beta_t^{task}]^\top [ M_{\lambda-1}] [ m_t^{task} \odot e^s].
\end{split}
\end{equation}

\begin{figure}[htbp]
    \centering
    \includegraphics[width=\linewidth]{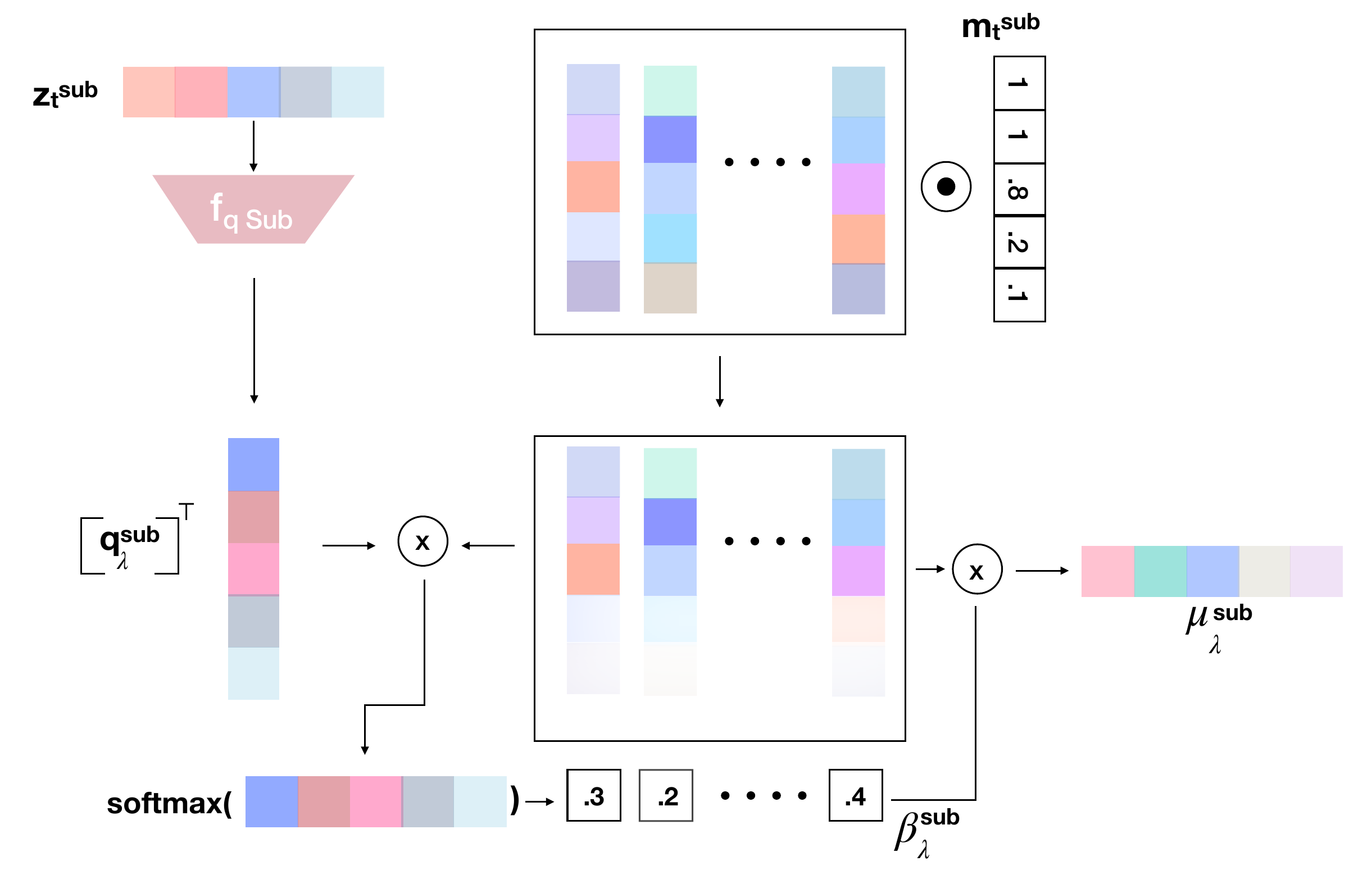}
    \caption{Illustration of Subject-Specific Feature Retrieval Using the Proposed Multi-Head Retrieval Scheme): Using the augmented mask, $m_t^{sub}$, which we generate for disengagement of the subject-specific features we first segment the relevant subject-specific memory content within each of the memory slots. Then using the factorised subject-specific feature vector, $z_t^{sub}$, we generate a subject-specific query vector, $q_{\lambda}^{sub}$, and we attend to the content of each of the slots and quantify the similarity between the content and $q_{\lambda}^{sub}$ which is captured by $\beta_{\lambda}^{sub}$. Finally, the subject-specific memory output, $\mu_{\lambda}^{sub}$, related to the subject-specific query vector is generated as per Eq. 12. }
    \label{fig:multi_head_access}
\end{figure}

Then, we aggregate the information retrieved using the overall, subject-specific, and task-specific queries and concatenate them across the final dimension, 
\begin{equation}
    \dot{\mu}_{\lambda} = [ \mu_{\lambda} \bigoplus \mu^{sub}_{\lambda} \bigoplus \mu^{task}_{\lambda}],
\end{equation}
where $\dot{\mu}_{\lambda} \in \mathbb{R}^{K_3 \times 3F}$. We also concatenate $\dot{\mu}_{\lambda}$ output with $z_t^{DN_3}$ such that $\Tilde{z_t} = [z_t^{DN_3} \bigoplus \dot{\mu}_{\lambda}]$ and $\Tilde{z_t} \in \mathbb{R}^{K_3 \times 4F}$. We apply average pooling across the final dimension of $\Tilde{z_t}$ and the resultant feature vector, $\ddot{z_t} \in \mathbb{R}^{K_3 \times F}$. This augmented feature vector is then used to predict future motion. Fig. \ref{fig:multi_head_access} visually illustrates the subject-specific feature retrieval process using the proposed multi-head retrieval scheme. Note that an identical process is leveraged to retrieve task-specific features from the auxiliary memory. 

The architecture of our decoder module which predicts future motion is presented in Sec. \ref{sec:decoder}, and in Sec.\ref{sec:memory_update} we outline the process of memory update which allows our model to perform continual learning. 

\subsection{Memory Update and Auxiliary Memory Stabilisation Losses}\label{sec:memory_update}
Once informative content is retrieved from the auxiliary memory the next task is to update the memory module to ensure the continual evolution of the knowledge embedded in  our memory. For this task the typical process is to leverage the memory output, $\dot{\mu}_{\lambda}$, and pass it through a non-linear function, $f_{WRITE}$ to generate a vector to update the memory such that,
\begin{equation}
    \Breve{\mu}_{\lambda} = f_{WRITE}(\dot{\mu}_{\lambda}),
\end{equation}
where $\Breve{\mu}_{\lambda} \in \mathbb{R}^{K_3 \times F}$.
Then the subsequent state of the memory can be generated using,
\begin{equation}
M_{\lambda} = M_{\lambda -1} [I - \beta_{\lambda} \odot e^F]^{\top} + [\Breve{\mu}_{\lambda} \odot e^s][\beta_{\lambda}\odot e^F]^{\top},
\end{equation}
where $I$ is a matrix of ones, $e^s \in \mathbb{R}^s$, $e^F \in \mathbb{R}^F$ are vectors of ones, $\odot$ denotes the outer product which duplicates its left vector $s$ or $F$ times to form a matrix and,
\begin{equation}
\beta_{\lambda} = \beta^{sub}_{\lambda} + \beta^{task}_{\lambda} + \beta^{DN_3}_{\lambda}.
\end{equation}

However, our prior investigations \cite{fernando2019memory, fernando2020detection, fernando2020neural} revealed that reliance on the downstream (i.e. classification, regression) objective alone could lead to sub-optimal memory states. For instance, imbalanced or poorly curated datasets could lead the memory to memorize the most frequently observed types of data, ignoring the less frequent classes as these contribute less to the overall loss. Moreover, if the diversity of the dataset is high, or if the embedding size or the number of memory slots is small, this could lead to frequent memory updates making the knowledge stored in the memory highly volatile and unstable making the discovery of long-term dependencies infeasible. To this end, we propose two novel architectural innovations. First, we incorporate an additional neural network-based predictor parameterized by the function $f_{\beta}$ which predicts the memory slots that need updating where,  
\begin{equation}
    \Breve{\beta}_{\lambda} = f_{\beta}(\Breve{m}_{\lambda}, \beta_{\lambda}).
    \label{eq:update_slot_prediction}
\end{equation}
Second, we propose two loss functions that directly operate on the memory content and encourage diversity and stability. Specifically, our diversity loss, $L_{div}$, will iteratively estimate the similarity between a slot's content and the rest of the content in memory. Formally, our diversity loss can be defined as,
\begin{equation}
    L_{div} = \frac{1}{s(s-1)} \sum_{i=1}^s\sum_{\substack{j=1 \\ i< j}}^{s}f_{COS} (M_{\lambda-1}^i, M_{\lambda-1}^j),
\end{equation}
where $f_{COS}$ is the cosine similarity loss and $M_{\lambda-1}^i$ denotes the $i^{th}$ memory slot in the memory. 

Next, we introduce $L_{cons}$, which helps consolidate memory content and penalises unstable memory updates. Formally, let $\Breve{\beta}^i_{\lambda}$ denote the $i^{th}$ element (i.e. $i^{th}$ memory slot) in the output of Eq. \ref{eq:update_slot_prediction} at time instance $\lambda$, let $w$ define the window size, then we generate $\frac{\lambda}{w}$ windows to inspect the change in memory updates for the period from time instance 1 to $\lambda$. The change in memory updates between two consecutive time intervals can be evaluated as,
\begin{equation}
    \Delta_j^i = |log\beta_j^i - log\beta_{j-1}^i|,
    \label{eq:memory_change}
\end{equation}
where,
\begin{equation}
    \Delta_c^i= 
\begin{cases}
    \frac{1}{|w|}\sum_{j=0}^w\Delta_j^i,& \text{if } \lambda-cw > j > \lambda-2cw\\
    0,              & \text{otherwise}
\end{cases}
\end{equation}

Now we define the memory consolidation loss, $L_{cons}$, as,
\begin{equation}
    L_{cons} = \frac{1}{|s|} \sum_{i \in s}\sum_{c \in \frac{\lambda}{w}} \Delta_c^i 2^{c}.
    \label{eq:loss_cons}
\end{equation}

This formulation of the loss penalises large memory updates that occur between large temporal intervals. Using the windowing operation in Eq. \ref{eq:memory_change}, we obtain the average amount of the change of the slot content within the window and using Eq. \ref{eq:loss_cons}, and penalise large changes (due to the log scale) that occur once the content has been initially written (due to the exponential scale of $c$). Using the window size we can control the period that we allow the slot content to stabilise and after which we exponentially penalise the content updates. As such, this loss does not negatively impact the continual learning ability of the memory. 

We highlight the use of $\Breve{\beta}_{\lambda}$ for the loss calculations instead of using $M_{\lambda-1}$ directly. Direct use of the memory content would require the individual memory states at all time instances to be stored, leading to inefficiencies. However, $\Breve{\beta}_{\lambda}$ provides a snapshot of how individual memory slots have been updated, and as such, we utilise this vector.

\subsection{Multi-Scale GCN Decoder}\label{sec:decoder}
Mirroring the downsampling blocks in our encoder, our decoder is equipped with upsampling GCN modules that gradually increase the resolution of the predicted pose representations. Specifically, the first upsampling block, $AN_3$, receives $\ddot{z_t} \in \mathbb{R}^{K_3 \times F}$ and decodes it to a feature vector $z_t^{AN_2} \in \mathbb{R}^{K_2 \times F}$. Similar to our encoder, our decoder has four upsampling blocks, $AN_3, AN_2, AN_1$, and $AN_0$. As in \cite{dang2021msr} four end-GCNs, $E_0, E_1, E_2$, and $E_3$, each with two graph convolution layers, are used to generate future pose sequences at four different scales. However, in contrast to \cite{dang2021msr} where respective downsampled (i.e. $z_t^{DN_0}, z_t^{DN_1}, z_t^{DN_2}, z_t^{DN_3}$) and upsampled (i.e. $z_t^{AN_0}, z_t^{AN_1}, z_t^{AN_2}, z_t^{AN_3}$) feature vectors in individual scales are concatenated and passed to the respective end GCNs, we only pass the upsampled features.

We use $L_2$ loss between the ground truth and predicted pose sequences to train the pose prediction head of our framework. We denote this loss as $L_{pose}$. 

The overall loss function of our framework then becomes 
\begin{multline}
    L^{*} = \theta_{pose}L_{pose} + \theta_{div}L_{div} + \\
     \theta_{cons}L_{cons} + \theta_{sub}L_{sub} + \theta_{task}L_{task},
\end{multline}
where $L_{sub}$ is the contrastive loss for the subject identification task and $L_{task}$ is the categorical cross-entropy loss for task identification. $\theta_{pose}, \theta_{div}, \theta_{cons}, \theta_{sub}$ and $\theta_{task}$ are loss weights that control the contributions from individual losses. 

\subsection{Implementation Details}\label{sec:implementation}
Implementation of this framework is completed using PyTorch. The Adam \cite{kingma2014adam} optimiser with an initial learning rate of $2e^{-4}$ is used for optimisation. The learning rate is decreased by $0.98$ every two epochs. The model is trained for 100 epochs on an NVIDIA A100 GPU. The embedding size, $F$, was experimentally chosen and was set to 300.  Similarly, hyper-parameters, $w, \theta_{pose}, \theta_{div}, \theta_{cons}, \theta_{sub}, \theta_{task}$ were experimentally chosen and were set to $15$, $0.4$, $0.15$, $0.15$, $0.15$, and $0.15$ respectively.

\section{Experiments}
In this section, we report the results of experiments that we conducted to evaluate and compare the efficiency of the proposed skeleton-based human motion prediction model, FMS-AM, with respect to existing state-of-the-art methods. We first introduce the details of the two datasets that we used for our evaluations (Sec. \ref{sec:datasets}), then present the evaluation metrics that we use to measure the model performance (Sec. \ref{sec:evalprotocol}). The main experimental results where we compare our proposed method with existing state-of-the-art approaches are presented in Sec. \ref{Sec:main_results}. Ablation evaluations that were conducted to demonstrate the efficacy of the proposed feature disentanglement strategy, the multi-head retrieval strategy, the novel stabilisation losses, and the dynamic mask generation process are presented in Sec. \ref{Sec:ablations}. In Sec. \ref{sec:time_complexity} we discuss the time complexity of our FMS-AM model.

\subsection{Datasets}
\label{sec:datasets}
For our evaluations, following the state-of-the-art methods we use two popular motion capture benchmark datasets, namely, the Human3.6M (H3.6M) and CMU Motion Capture (CMU-Mocap) datasets. Details of these datasets are provided in the following subsections.

\subsubsection{H3.6M dataset \cite{ionescu2013human3}} The H3.6M dataset consists of motions that performed by 11 professional actors, 5 female, and 6 male. This is a challenging dataset with 15 different action categories, including, Taking Photos, Waiting, Giving Directions, Walking Pair, Phone Talk, Sitting on the floor, Smoking, Sitting on a chair, etc. Similar to prior works \cite{mao2019learning, li2020dynamic, dang2021msr} we use the data of seven subjects, S1, S5, S6, S7, S8, S9, and S11. Following \cite{mao2019learning, li2020dynamic, dang2021msr} we use the data from S5 for testing, S11 for validation, and the rest of the subjects for model training. 22 body joints from the original 32 joints are chosen to represent the body pose and the data is mapped to a 3D joint coordinate space. We downsample all pose sequences by a factor of two along the temporal axis. 
    
\subsubsection{CMU-Mocap dataset} The CMU-Mocap dataset \footnote{http://mocap.cs.cmu.edu/} has 5 abstract action classes including `human interaction', `interaction with environment', `locomotion', `physical activities \& sports', and `situations \& scenarios'. To maintain consistency with the H3.6M dataset we choose the following 8 detailed action categories from the dataset: basketball, basketball signal, directing traffic, jumping, running, soccer, walking, and washing a window. Similar to the H3.6M dataset 22 body joints from the original 38 joints were filtered and the data is mapped to 
a 3D joint coordinate space. 

\subsection{Evaluation Protocol}
\label{sec:evalprotocol}

Mean Per Joint Position Error (MPJPE) has been widely used as the evaluation metric in numerous recent works \cite{dang2021msr, ma2022progressively, tang2023collaborative} due to its ability to directly compare different frameworks, its intuitive nature, and the ability to evaluate it directly on
skeleton kinematics. Therefore, as our evaluation metric we report MPJPE in millimeters, calculated using
\begin{equation}
    L_{\mathrm{MPJPE}} = \frac{1}{J \times T} \sum_{t=1}^T \sum_{j=1}^J ||\hat{p}_{j,t} - p_{j,t}||^2,
\end{equation}
where $\hat{p}_{j,t} \in \mathbb{R}^3$ is the predicted position of $j^{th}$ joint in $t^{th}$ frame while $p_{j,t}$ is the corresponding ground truth. Lower error values indicate better agreement between the predictions and ground truths. 

As in \cite{li2020dynamic, dang2021msr} we generate predictions for different short-term and long-term prediction horizons. Specifically, pose sequence predictions of 80 ms, 160 ms, 320 ms, and 400 ms (i.e. 10 frame sequence) were generated as short-term predictions, while 560ms and 1000ms (i.e. 25 frames) length sequences were generated for long-term predictions.

\subsection{Comparisons with Existing State-of-the-art Methods}\label{Sec:main_results}

As baseline methods, we use state-of-the-art methods including Residual Sup \cite{martinez2017human}, Traj-GCN \cite{mao2019learning}, DMGNN \cite{li2020dynamic}, MSR-GCN \cite{dang2021msr}, S-DGCN \cite{ma2022progressively}, PK-GCN \cite{sun2022overlooked}, DANet \cite{cao2022dual}, DPnet \cite{tang2023collaborative} and GA-MIN \cite{zhong2023geometric}. When choosing our baselines we ensured that a variety of different deep learning approaches, including recurrent neural networks, attention-based methods, graph neural networks, hybrid approaches, multi-scale and multi-stage GCN architecture, and geometric-inspired neural network architectures, are compared, enabling a comprehensive comparison. 

Quantitative comparisons for short-term and long-term prediction results for the H3.6M dataset are presented in Tabs. \ref{tab:H3.6M_short_term} and \ref{tab:H3.6M_long_term}, respectively. Following \cite{sun2022overlooked, li2020dynamic, zhou2021learning,dang2021msr} for long-term prediction we report average error metrics and evaluations only for five popular action categories for ease of presentation. GCN-based approaches such as MSR-GCN \cite{dang2021msr}, S-DGCN \cite{ma2022progressively}, PK-GCN \cite{sun2022overlooked} have been able to achieve comparatively higher performance compared to the RNN-based Residual sup method. However, demonstrating the feature contamination issue when aggregating high-dimensional features using multi-stage graph convolution operation, they struggle to reach the robustness level of GA-MIN \cite{zhong2023geometric}. However, when comparing FMS-AM with GA-MIN, which is the previous state-of-the-art method we observe that our method has the lowest MPJPE on average for both short-term and long-term predictions. Specifically,  we observe that the GA-MIN method struggles to generate accurate pose predictions over 400 ms on the H3.6M dataset. Large errors in the long-term prediction setting (i.e. Tab. \ref{tab:H3.6M_long_term}, 1000ms setting) for classes such as directions (i.e MPJPE $\sim$ 100) and discussion (i.e MPJPE $>$ 106) were observed where there is subject specificity compared to more generalised actions such as walking where we observe comparatively lower MPJPE (i.e MPJPE $\sim$ 43). In contrast, our FMS-AM framework has been able to achieve consistent performance across all the action classes irrespective of subject-specific or class-specific motions. The average results of MPJPEs from Tabs \ref{tab:H3.6M_short_term} and \ref{tab:H3.6M_long_term} show that FMS-AM on average lowers the short-term prediction error (at 400ms) by 17.40\% and the long-term prediction error by 17.54\%, both significant margins. We believe the proposed feature factorisation strategy coupled with the innovative multi-head retrieval of the auxiliary memory has allowed superior learning capabilities where subject-specific, class-specific, and auxiliary information are better incorporated into the prediction of future motion sequences. 

\begin{table*}[htbp]
\caption{Comparisons between the proposed method and state-of-the-art methods in terms of MPJPE for short-term prediction on the 15 action categories of H3.6M dataset. We also report the average across the action categories. The best results are highlighted in bold}
\label{tab:H3.6M_short_term}
\resizebox{\textwidth}{!}{%
\begin{tabular}{|c|cccc|cccc|cccc|cccc|}
\hline
\multirow{2}{*}{Model} & \multicolumn{4}{c|}{walking}                                                                                 & \multicolumn{4}{c|}{eating}                                                                                  & \multicolumn{4}{c|}{smoking}                                                                                 & \multicolumn{4}{c|}{discussion}                                                                          \\ \cline{2-17} 
                       & \multicolumn{1}{c|}{80}        & \multicolumn{1}{c|}{160}       & \multicolumn{1}{c|}{320}       & 400       & \multicolumn{1}{c|}{80}        & \multicolumn{1}{c|}{160}       & \multicolumn{1}{c|}{320}       & 400       & \multicolumn{1}{c|}{80}        & \multicolumn{1}{c|}{160}       & \multicolumn{1}{c|}{320}       & 400       & \multicolumn{1}{c|}{80}        & \multicolumn{1}{c|}{160}       & \multicolumn{1}{c|}{320}   & 400       \\ \hline
Residual Sup \cite{martinez2017human}          & \multicolumn{1}{c|}{29.36}     & \multicolumn{1}{c|}{50.82}     & \multicolumn{1}{c|}{76.03}     & 81.51     & \multicolumn{1}{c|}{16.84}     &  \multicolumn{1}{c|}{30.60}     & \multicolumn{1}{c|}{56.92}     & 68.65     & \multicolumn{1}{c|}{22.96}     & \multicolumn{1}{c|}{42.64}     & \multicolumn{1}{c|}{70.14}     & 82.68     & \multicolumn{1}{c|}{32.94}     & \multicolumn{1}{c|}{61.18}     & \multicolumn{1}{c|}{90.92} & 96.19     \\ \hline
DMGNN    \cite{li2020dynamic}              & \multicolumn{1}{c|}{17.32}     & \multicolumn{1}{c|}{30.67}     & \multicolumn{1}{c|}{54.56}     & 65.20     & \multicolumn{1}{c|}{10.96}     & \multicolumn{1}{c|}{21.39}     & \multicolumn{1}{c|}{36.18}     & 43.88     & \multicolumn{1}{c|}{8.97}      & \multicolumn{1}{c|}{17.62}     & \multicolumn{1}{c|}{32.05}     & 40.30     & \multicolumn{1}{c|}{17.33}     & \multicolumn{1}{c|}{34.78}     & \multicolumn{1}{c|}{61.03} & 69.80     \\ \hline
Traj-GCN  \cite{mao2019learning}             & \multicolumn{1}{c|}{12.29}     & \multicolumn{1}{c|}{23.03}     & \multicolumn{1}{c|}{39.77}     & 46.12     & \multicolumn{1}{c|}{8.36}      & \multicolumn{1}{c|}{16.90}     & \multicolumn{1}{c|}{33.19}     & 40.70     & \multicolumn{1}{c|}{7.94}      & \multicolumn{1}{c|}{16.24}     & \multicolumn{1}{c|}{31.90}     & 38.90     & \multicolumn{1}{c|}{12.50}     & \multicolumn{1}{c|}{27.40}     & \multicolumn{1}{c|}{58.51} & 71.68     \\ \hline
MSR-GCN \cite{dang2021msr}               & \multicolumn{1}{c|}{12.16}     & \multicolumn{1}{c|}{22.65}     & \multicolumn{1}{c|}{38.64}     & 45.24     & \multicolumn{1}{c|}{8.39}      & \multicolumn{1}{c|}{17.05}     & \multicolumn{1}{c|}{33.03}     & 40.43     & \multicolumn{1}{c|}{8.02}      & \multicolumn{1}{c|}{16.27}     & \multicolumn{1}{c|}{31.32}     & 38.15     & \multicolumn{1}{c|}{11.98}     & \multicolumn{1}{c|}{26.76}     & \multicolumn{1}{c|}{57.08} & 69.74     \\ \hline
S-DGCN \cite{ma2022progressively}     & \multicolumn{1}{c|}{9.5}       & \multicolumn{1}{c|}{19.7}      & \multicolumn{1}{c|}{34.6}      & 40.0      & \multicolumn{1}{c|}{7.7}       & \multicolumn{1}{c|}{16.2}      & \multicolumn{1}{c|}{31.1}      & 38.9      & \multicolumn{1}{c|}{6.8}       & \multicolumn{1}{c|}{14.5}      & \multicolumn{1}{c|}{28.1}      & 34.9      & \multicolumn{1}{c|}{9.1}       & \multicolumn{1}{c|}{20.6}      & \multicolumn{1}{c|}{52.4}  & 66.2      \\ \hline
PK-GCN \cite{sun2022overlooked}                & \multicolumn{1}{c|}{8.9}       & \multicolumn{1}{c|}{15.9}      & \multicolumn{1}{c|}{28.0}      & 31.6      & \multicolumn{1}{c|}{8.1}       & \multicolumn{1}{c|}{17.7}      & \multicolumn{1}{c|}{33.6}      & 41.8      & \multicolumn{1}{c|}{7.4}       & \multicolumn{1}{c|}{14.3}      & \multicolumn{1}{c|}{24.4}      & 29.2      & \multicolumn{1}{c|}{10.3}      & \multicolumn{1}{c|}{22.9}      & \multicolumn{1}{c|}{42.0}  & 47.2      \\ \hline
DANet \cite{cao2022dual}      & \multicolumn{1}{c|}{9.7}       & \multicolumn{1}{c|}{19.0}      & \multicolumn{1}{c|}{33.5}      & 39.4      & \multicolumn{1}{c|}{6.1}       & \multicolumn{1}{c|}{13.6}      & \multicolumn{1}{c|}{27.6}      & 34.7      & \multicolumn{1}{c|}{6.4}       & \multicolumn{1}{c|}{13.1}      & \multicolumn{1}{c|}{25.6}      & 31.9      & \multicolumn{1}{c|}{8.8}       & \multicolumn{1}{c|}{19.1}      & \multicolumn{1}{c|}{39.6}  & 50.1      \\ \hline
DPnet \cite{tang2023collaborative}                 & \multicolumn{1}{c|}{7.3}       & \multicolumn{1}{c|}{15.2}      & \multicolumn{1}{c|}{30.1}      & 32.6      & \multicolumn{1}{c|}{8.6}       & \multicolumn{1}{c|}{18.3}      & \multicolumn{1}{c|}{36.4}      & 43.5      & \multicolumn{1}{c|}{6.9}       & \multicolumn{1}{c|}{13.5}      & \multicolumn{1}{c|}{24.3}      & 28.7      & \multicolumn{1}{c|}{8.2}       & \multicolumn{1}{c|}{20.1}      & \multicolumn{1}{c|}{38.2}  & 43.0      \\ \hline
GA-MIN \cite{zhong2023geometric}  & \multicolumn{1}{c|}{7.5}       & \multicolumn{1}{c|}{13.5}      & \multicolumn{1}{c|}{28.2}      & 30.6      & \multicolumn{1}{c|}{5.8}       & \multicolumn{1}{c|}{12.5}      & \multicolumn{1}{c|}{25.3}      & 33.8      & \multicolumn{1}{c|}{6.2}       & \multicolumn{1}{c|}{12.1}      & \multicolumn{1}{c|}{24.2}      & 24.2      & \multicolumn{1}{c|}{8.2}       & \multicolumn{1}{c|}{18.6}      & \multicolumn{1}{c|}{30.5}  & 46.3      \\ \hline
FMS-AM                   & \multicolumn{1}{c|}{\textbf{5.1}} & \multicolumn{1}{c|}{\textbf{10.3}} & \multicolumn{1}{c|}{\textbf{18.8}} & \textbf{20.4} & \multicolumn{1}{c|}{\textbf{4.2}} & \multicolumn{1}{c|}{\textbf{8.7}} & \multicolumn{1}{c|}{\textbf{16.5}} & \textbf{21.2} & \multicolumn{1}{c|}{\textbf{4.6}} & \multicolumn{1}{c|}{\textbf{9.3}} & \multicolumn{1}{c|}{\textbf{17.1}} & \textbf{12.7} & \multicolumn{1}{c|}{\textbf{5.5}} & \multicolumn{1}{c|}{\textbf{12.1}} & \multicolumn{1}{c|}{\textbf{19.6}}      & \textbf{26.8} \\ \hline

\hline
\hline
\multirow{2}{*}{Model}             & \multicolumn{4}{c|}{directions}                                                                                               & \multicolumn{4}{c|}{greeting}                                                                                                  & \multicolumn{4}{c|}{phoning}                                                                                                   & \multicolumn{4}{c|}{posing}                                                                                                 \\ \cline{2-17} 
                                   & \multicolumn{1}{c|}{80}        & \multicolumn{1}{c|}{160}       & \multicolumn{1}{c|}{320}       & 400                        & \multicolumn{1}{c|}{80}        & \multicolumn{1}{c|}{160}       & \multicolumn{1}{c|}{320}       & 400                         & \multicolumn{1}{c|}{80}        & \multicolumn{1}{c|}{160}       & \multicolumn{1}{c|}{320}       & 400                         & \multicolumn{1}{c|}{80}        & \multicolumn{1}{c|}{160}       & \multicolumn{1}{c|}{320}    & 400                         \\ \hline
\multicolumn{1}{|l|}{Residual Sup \cite{martinez2017human}} & \multicolumn{1}{l|}{35.36}     & \multicolumn{1}{l|}{57.27}     & \multicolumn{1}{l|}{76.30}     & \multicolumn{1}{l|}{87.67} & \multicolumn{1}{l|}{34.46}     & \multicolumn{1}{l|}{63.36}     & \multicolumn{1}{l|}{124.60}    & \multicolumn{1}{l|}{142.50} & \multicolumn{1}{l|}{37.96}     & \multicolumn{1}{l|}{69.32}     & \multicolumn{1}{l|}{115.00}    & \multicolumn{1}{l|}{126.73} & \multicolumn{1}{l|}{36.10}     & \multicolumn{1}{l|}{69.12}     & \multicolumn{1}{l|}{130.46} & \multicolumn{1}{l|}{157.08} \\ \hline
DMGNN   \cite{li2020dynamic}                           & \multicolumn{1}{c|}{13.14}     & \multicolumn{1}{c|}{24.62}     & \multicolumn{1}{c|}{64.68}     & 81.86                      & \multicolumn{1}{c|}{23.30}     & \multicolumn{1}{c|}{50.32}     & \multicolumn{1}{c|}{107.30}    & 132.10                      & \multicolumn{1}{c|}{12.47}     & \multicolumn{1}{c|}{25.77}     & \multicolumn{1}{c|}{48.08}     & 58.29                       & \multicolumn{1}{c|}{15.27}     & \multicolumn{1}{c|}{29.27}     & \multicolumn{1}{c|}{71.54}  & 96.65                       \\ \hline
Traj-GCN \cite{mao2019learning}                          & \multicolumn{1}{c|}{8.97}      & \multicolumn{1}{c|}{19.87}     & \multicolumn{1}{c|}{43.35}     & 53.74                      & \multicolumn{1}{c|}{18.65}     & \multicolumn{1}{c|}{38.68}     & \multicolumn{1}{c|}{77.74}     & 93.39                       & \multicolumn{1}{c|}{10.24}     & \multicolumn{1}{c|}{21.02}     & \multicolumn{1}{c|}{42.54}     & 52.30                       & \multicolumn{1}{c|}{13.66}     & \multicolumn{1}{c|}{29.89}     & \multicolumn{1}{c|}{66.62}  & 84.05                       \\ \hline
MSR-GCN \cite{dang2021msr}                           & \multicolumn{1}{c|}{8.61}      & \multicolumn{1}{c|}{19.65}     & \multicolumn{1}{c|}{43.28}     & 53.82                      & \multicolumn{1}{c|}{16.48}     & \multicolumn{1}{c|}{36.95}     & \multicolumn{1}{c|}{77.32}     & 93.38                       & \multicolumn{1}{c|}{10.10}     & \multicolumn{1}{c|}{20.74}     & \multicolumn{1}{c|}{41.51}     & 51.26                       & \multicolumn{1}{c|}{12.79}     & \multicolumn{1}{c|}{29.38}     & \multicolumn{1}{c|}{66.95}  & 85.01                       \\ \hline
S-DGCN \cite{ma2022progressively}                 & \multicolumn{1}{c|}{8.3}       & \multicolumn{1}{c|}{18.8}      & \multicolumn{1}{c|}{42.5}      & 51.9                       & \multicolumn{1}{c|}{13.7}      & \multicolumn{1}{c|}{30.4}      & \multicolumn{1}{c|}{68.6}      & 85.3                        & \multicolumn{1}{c|}{8.1}       & \multicolumn{1}{c|}{18.0}      & \multicolumn{1}{c|}{37.6}      & 47.9                        & \multicolumn{1}{c|}{8.8}       & \multicolumn{1}{c|}{22.9}      & \multicolumn{1}{c|}{58.3}   & 73.8                        \\ \hline
PK-GCN \cite{sun2022overlooked}                            & \multicolumn{1}{c|}{8.6}       & \multicolumn{1}{c|}{23.7}      & \multicolumn{1}{c|}{46.5}      & 56.2                       & \multicolumn{1}{c|}{13.3}      & \multicolumn{1}{c|}{27.2}      & \multicolumn{1}{c|}{67.3}      & 83.1                        & \multicolumn{1}{c|}{11.4}      & \multicolumn{1}{c|}{20.2}      & \multicolumn{1}{c|}{37.7}      & 43.2                        & \multicolumn{1}{c|}{9.1}       & \multicolumn{1}{c|}{23.6}      & \multicolumn{1}{c|}{65.8}   & 81.2                        \\ \hline
DANet \cite{cao2022dual}                  & \multicolumn{1}{c|}{6.9}       & \multicolumn{1}{c|}{17.6}      & \multicolumn{1}{c|}{43.0}      & 54.9                       & \multicolumn{1}{c|}{12.7}      & \multicolumn{1}{c|}{28.6}      & \multicolumn{1}{c|}{61.6}      & 75.9                        & \multicolumn{1}{c|}{8.3}       & \multicolumn{1}{c|}{17.9}      & \multicolumn{1}{c|}{38.2}      & 48.3                        & \multicolumn{1}{c|}{8.4}       & \multicolumn{1}{c|}{19.4}      & \multicolumn{1}{c|}{43.4}   & 56.1                        \\ \hline
DPnet   \cite{tang2023collaborative}                           & \multicolumn{1}{c|}{10.1}      & \multicolumn{1}{c|}{21.0}      & \multicolumn{1}{c|}{45.8}      & 56.7                       & \multicolumn{1}{c|}{12.7}      & \multicolumn{1}{c|}{27.1}      & \multicolumn{1}{c|}{65.6}      & 82.9                        & \multicolumn{1}{c|}{10.2}      & \multicolumn{1}{c|}{17.4}      & \multicolumn{1}{c|}{35.7}      & 41.3                        & \multicolumn{1}{c|}{7.4}       & \multicolumn{1}{c|}{21.9}      & \multicolumn{1}{c|}{63.5}   & 78.8                        \\ \hline
GA-MIN \cite{zhong2023geometric}              & \multicolumn{1}{c|}{6.8}       & \multicolumn{1}{c|}{15.3}      & \multicolumn{1}{c|}{42.1}      & 50.2                       & \multicolumn{1}{c|}{12.8}      & \multicolumn{1}{c|}{26.3}      & \multicolumn{1}{c|}{61.8}      & 75.8                        & \multicolumn{1}{c|}{8.3}       & \multicolumn{1}{c|}{17.8}      & \multicolumn{1}{c|}{37.9}      & 44.8                        & \multicolumn{1}{c|}{7.8}       & \multicolumn{1}{c|}{19.3}      & \multicolumn{1}{c|}{43.4}   & 56.0                        \\ \hline
FMS-AM                               & \multicolumn{1}{c|}{\textbf{4.3}} & \multicolumn{1}{c|}{\textbf{10.1}} & \multicolumn{1}{c|}{\textbf{25.4}} & \textbf{30.1}                  & \multicolumn{1}{c|}{\textbf{9.1}} & \multicolumn{1}{c|}{\textbf{18.2}} & \multicolumn{1}{c|}{\textbf{42.4}} & \textbf{53.7}                   & \multicolumn{1}{c|}{\textbf{6.1}} & \multicolumn{1}{c|}{\textbf{10.2}} & \multicolumn{1}{c|}{\textbf{22.7}} & \textbf{28.4}                   & \multicolumn{1}{c|}{\textbf{5.4}} & \multicolumn{1}{c|}{\textbf{14.1}} & \multicolumn{1}{c|}{\textbf{19.8} }      & \textbf{33.4}                   \\ \hline

\hline
\hline
\multirow{2}{*}{Model}             & \multicolumn{4}{c|}{purchases}                                                                                                & \multicolumn{4}{c|}{sitting}                                                                                                   & \multicolumn{4}{c|}{sittingdown}                                                                                               & \multicolumn{4}{c|}{takingphoto}                                                                                          \\ \cline{2-17} 
                                   & \multicolumn{1}{c|}{80}        & \multicolumn{1}{c|}{160}       & \multicolumn{1}{c|}{320}       & 400                        & \multicolumn{1}{c|}{80}        & \multicolumn{1}{c|}{160}       & \multicolumn{1}{c|}{320}       & 400                         & \multicolumn{1}{c|}{80}        & \multicolumn{1}{c|}{160}       & \multicolumn{1}{c|}{320}       & 400                         & \multicolumn{1}{c|}{80}        & \multicolumn{1}{c|}{160}       & \multicolumn{1}{c|}{320}   & 400                        \\ \hline
\multicolumn{1}{|l|}{Residual Sup \cite{martinez2017human}} & \multicolumn{1}{l|}{36.33}     & \multicolumn{1}{l|}{60.30}     & \multicolumn{1}{l|}{86.53}     & \multicolumn{1}{l|}{95.92} & \multicolumn{1}{l|}{42.55}     & \multicolumn{1}{l|}{81.40}     & \multicolumn{1}{l|}{134.70}    & \multicolumn{1}{l|}{151.78} & \multicolumn{1}{l|}{47.28}     & \multicolumn{1}{l|}{85.95}     & \multicolumn{1}{l|}{145.75}    & \multicolumn{1}{l|}{168.86} & \multicolumn{1}{l|}{26.10}     & \multicolumn{1}{l|}{47.61}     & \multicolumn{1}{l|}{81.40} & \multicolumn{1}{l|}{94.73} \\ \hline
DMGNN \cite{li2020dynamic}                             & \multicolumn{1}{c|}{21.35}     & \multicolumn{1}{c|}{38.71}     & \multicolumn{1}{c|}{75.67}     & 92.74                      & \multicolumn{1}{c|}{11.92}     & \multicolumn{1}{c|}{25.11}     & \multicolumn{1}{c|}{44.59}     & 50.20                       & \multicolumn{1}{c|}{14.95}     & \multicolumn{1}{c|}{32.88}     & \multicolumn{1}{c|}{77.06}     & 93.00                       & \multicolumn{1}{c|}{13.61}     & \multicolumn{1}{c|}{28.95}     & \multicolumn{1}{c|}{45.99} & 58.79                      \\ \hline
Traj-GCN \cite{mao2019learning}                          & \multicolumn{1}{c|}{15.60}     & \multicolumn{1}{c|}{32.78}     & \multicolumn{1}{c|}{65.72}     & 79.25                      & \multicolumn{1}{c|}{10.62}     & \multicolumn{1}{c|}{21.90}     & \multicolumn{1}{c|}{46.33}     & 57.91                       & \multicolumn{1}{c|}{16.14}     & \multicolumn{1}{c|}{31.12}     & \multicolumn{1}{c|}{61.47}     & 75.46                       & \multicolumn{1}{c|}{9.88}      & \multicolumn{1}{c|}{20.89}     & \multicolumn{1}{c|}{44.95} & 56.58                      \\ \hline
MSR-GCN  \cite{dang2021msr}                          & \multicolumn{1}{c|}{14.75}     & \multicolumn{1}{c|}{32.39}     & \multicolumn{1}{c|}{66.13}     & 79.64                      & \multicolumn{1}{c|}{10.53}     & \multicolumn{1}{c|}{21.99}     & \multicolumn{1}{c|}{46.26}     & 57.80                       & \multicolumn{1}{c|}{16.10}     & \multicolumn{1}{c|}{31.63}     & \multicolumn{1}{c|}{62.45}     & 76.84                       & \multicolumn{1}{c|}{9.89}      & \multicolumn{1}{c|}{21.01}     & \multicolumn{1}{c|}{44.56} & 56.30                      \\ \hline
S-DGCN \cite{ma2022progressively}                 & \multicolumn{1}{c|}{12.2}      & \multicolumn{1}{c|}{28.6}      & \multicolumn{1}{c|}{59.9}      & 74.3                       & \multicolumn{1}{c|}{9.0}       & \multicolumn{1}{c|}{20.0}      & \multicolumn{1}{c|}{43.5}      & 54.9                        & \multicolumn{1}{c|}{12.7}      & \multicolumn{1}{c|}{28.5}      & \multicolumn{1}{c|}{58.8}      & 72.3                        & \multicolumn{1}{c|}{8.3}       & \multicolumn{1}{c|}{18.6}      & \multicolumn{1}{c|}{43.1}  & 53.7                       \\ \hline
PK-GCN  \cite{sun2022overlooked}                           & \multicolumn{1}{c|}{15.2}      & \multicolumn{1}{c|}{31.4}      & \multicolumn{1}{c|}{57.9}      & 68.0                       & \multicolumn{1}{c|}{10.1}      & \multicolumn{1}{c|}{24.6}      & \multicolumn{1}{c|}{47.8}      & 57.3                        & \multicolumn{1}{c|}{11.5}      & \multicolumn{1}{c|}{27.5}      & \multicolumn{1}{c|}{56.8}      & 67.3                        & \multicolumn{1}{c|}{7.6}       & \multicolumn{1}{c|}{16.1}      & \multicolumn{1}{c|}{39.7}  & 51.3                       \\ \hline
DANet \cite{cao2022dual}                 & \multicolumn{1}{c|}{12.4}      & \multicolumn{1}{c|}{28.5}      & \multicolumn{1}{c|}{60.1}      & 73.6                       & \multicolumn{1}{c|}{8.9}       & \multicolumn{1}{c|}{19.3}      & \multicolumn{1}{c|}{42.6}      & 53.9                        & \multicolumn{1}{c|}{14.6}      & \multicolumn{1}{c|}{30.7}      & \multicolumn{1}{c|}{59.9}      & 73.0                        & \multicolumn{1}{c|}{8.0}       & \multicolumn{1}{c|}{17.7}      & \multicolumn{1}{c|}{39.9}  & 51.0                       \\ \hline
DPnet  \cite{tang2023collaborative}                            & \multicolumn{1}{c|}{17.8}      & \multicolumn{1}{c|}{37.0}      & \multicolumn{1}{c|}{62.1}      & 65.6                       & \multicolumn{1}{c|}{9.1}       & \multicolumn{1}{c|}{23.0}      & \multicolumn{1}{c|}{48.1}      & 62.8                        & \multicolumn{1}{c|}{9.7}       & \multicolumn{1}{c|}{24.2}      & \multicolumn{1}{c|}{49.7}      & 62.0                        & \multicolumn{1}{c|}{5.7}       & \multicolumn{1}{c|}{14.4}      & \multicolumn{1}{c|}{35.6}  & 47.9                       \\ \hline
GA-MIN \cite{zhong2023geometric}              & \multicolumn{1}{c|}{12.4}      & \multicolumn{1}{c|}{28.5}      & \multicolumn{1}{c|}{60.0}      & 72.9                       & \multicolumn{1}{c|}{8.1}       & \multicolumn{1}{c|}{18.5}      & \multicolumn{1}{c|}{41.9}      & 53.2                        & \multicolumn{1}{c|}{14.5}      & \multicolumn{1}{c|}{25.5}      & \multicolumn{1}{c|}{56.3}      & 70.3                        & \multicolumn{1}{c|}{8.3}       & \multicolumn{1}{c|}{16.6}      & \multicolumn{1}{c|}{38.2}  & 49.0                       \\ \hline
FMS-AM                               & \multicolumn{1}{c|}{\textbf{10.2}} & \multicolumn{1}{c|}{\textbf{20.8}} & \multicolumn{1}{c|}{\textbf{37.3}} & \textbf{51.5}                  & \multicolumn{1}{c|}{\textbf{5.9}} & \multicolumn{1}{c|}{\textbf{12.6}} & \multicolumn{1}{c|}{\textbf{20.8}} & \textbf{32.1}                   & \multicolumn{1}{c|}{\textbf{11.2}} & \multicolumn{1}{c|}{\textbf{17.1}} & \multicolumn{1}{c|}{\textbf{34.6}} & \textbf{46.4}                   & \multicolumn{1}{c|}{\textbf{6.1}} & \multicolumn{1}{c|}{\textbf{10.5}} & \multicolumn{1}{c|}{\textbf{27.0}}      & \textbf{36.2}                  \\ \hline \hline

\multirow{2}{*}{Model}             & \multicolumn{4}{c|}{waiting}                                                                                                   & \multicolumn{4}{c|}{walkingdog}                                                                                                & \multicolumn{4}{c|}{walkingtogether}                                                                                          & \multicolumn{4}{c|}{Average}                                                                                                \\ \cline{2-17} 
                                   & \multicolumn{1}{c|}{80}        & \multicolumn{1}{c|}{160}       & \multicolumn{1}{c|}{320}       & 400                         & \multicolumn{1}{c|}{80}        & \multicolumn{1}{c|}{160}       & \multicolumn{1}{c|}{320}       & 400                         & \multicolumn{1}{c|}{80}        & \multicolumn{1}{c|}{160}       & \multicolumn{1}{c|}{320}       & 400                        & \multicolumn{1}{c|}{80}        & \multicolumn{1}{c|}{160}       & \multicolumn{1}{c|}{320}    & 400                         \\ \hline
\multicolumn{1}{|l|}{Residual Sup \cite{martinez2017human}} & \multicolumn{1}{l|}{30.62}     & \multicolumn{1}{l|}{57.82}     & \multicolumn{1}{l|}{106.22}    & \multicolumn{1}{l|}{121.45} & \multicolumn{1}{l|}{64.18}     & \multicolumn{1}{l|}{102.10}    & \multicolumn{1}{l|}{141.07}    & \multicolumn{1}{l|}{164.35} & \multicolumn{1}{l|}{26.79}     & \multicolumn{1}{l|}{50.07}     & \multicolumn{1}{l|}{80.16}     & \multicolumn{1}{l|}{92.23} & \multicolumn{1}{l|}{34.66}     & \multicolumn{1}{l|}{61.97}     & \multicolumn{1}{l|}{101.08} & \multicolumn{1}{l|}{115.49} \\ \hline
DMGNN   \cite{li2020dynamic}                           & \multicolumn{1}{c|}{12.20}     & \multicolumn{1}{c|}{24.17}     & \multicolumn{1}{c|}{59.62}     & 77.54                       & \multicolumn{1}{c|}{47.09}     & \multicolumn{1}{c|}{93.33}     & \multicolumn{1}{c|}{160.13}    & 171.20                      & \multicolumn{1}{c|}{14.34}     & \multicolumn{1}{c|}{26.67}     & \multicolumn{1}{c|}{50.08}     & 63.22                      & \multicolumn{1}{c|}{16.95}     & \multicolumn{1}{c|}{33.62}     & \multicolumn{1}{c|}{65.90}  & 79.65                       \\ \hline
Traj-GCN  \cite{mao2019learning}                         & \multicolumn{1}{c|}{11.43}     & \multicolumn{1}{c|}{23.99}     & \multicolumn{1}{c|}{50.06}     & 61.48                       & \multicolumn{1}{c|}{23.39}     & \multicolumn{1}{c|}{46.17}     & \multicolumn{1}{c|}{83.47}     & 95.96                       & \multicolumn{1}{c|}{16.14}     & \multicolumn{1}{c|}{10.47}     & \multicolumn{1}{c|}{21.04}     & 38.47                      & \multicolumn{1}{c|}{12.68}     & \multicolumn{1}{c|}{26.06}     & \multicolumn{1}{c|}{52.27}  & 63.51                       \\ \hline
MSR-GCN  \cite{dang2021msr}                          & \multicolumn{1}{c|}{10.68}     & \multicolumn{1}{c|}{23.06}     & \multicolumn{1}{c|}{48.25}     & 59.23                       & \multicolumn{1}{c|}{20.65}     & \multicolumn{1}{c|}{42.88}     & \multicolumn{1}{c|}{80.35}     & 93.31                       & \multicolumn{1}{c|}{10.56}     & \multicolumn{1}{c|}{20.92}     & \multicolumn{1}{c|}{37.40}     & 43.85                      & \multicolumn{1}{c|}{12.11}     & \multicolumn{1}{c|}{25.56}     & \multicolumn{1}{c|}{51.64}  & 62.93                       \\ \hline
S-DGCN \cite{ma2022progressively}                 & \multicolumn{1}{c|}{8.7}       & \multicolumn{1}{c|}{19.4}      & \multicolumn{1}{c|}{43.7}      & 54.8                        & \multicolumn{1}{c|}{18.5}      & \multicolumn{1}{c|}{38.8}      & \multicolumn{1}{c|}{71.8}      & 85.3                        & \multicolumn{1}{c|}{8.5}       & \multicolumn{1}{c|}{18.3}      & \multicolumn{1}{c|}{35.2}      & 41.9                       & \multicolumn{1}{c|}{10.0}      & \multicolumn{1}{c|}{22.2}      & \multicolumn{1}{c|}{47.3}   & 58.4                        \\ \hline
PK-GCN  \cite{sun2022overlooked}                           & \multicolumn{1}{c|}{9.5}       & \multicolumn{1}{c|}{23.0}      & \multicolumn{1}{c|}{55.9}      & 63.6                        & \multicolumn{1}{c|}{21.3}      & \multicolumn{1}{c|}{42.4}      & \multicolumn{1}{c|}{83.7}      & 95.1                        & \multicolumn{1}{c|}{9.4}       & \multicolumn{1}{c|}{19.3}      & \multicolumn{1}{c|}{36.3}      & 44.8                       & \multicolumn{1}{c|}{10.8}      & \multicolumn{1}{c|}{23.3}      & \multicolumn{1}{c|}{48.2}   & 57.4                        \\ \hline
DANet \cite{cao2022dual}                  & \multicolumn{1}{c|}{8.1}       & \multicolumn{1}{c|}{18.3}      & \multicolumn{1}{c|}{41.6}      & 52.8                        & \multicolumn{1}{c|}{19.0}      & \multicolumn{1}{c|}{38.7}      & \multicolumn{1}{c|}{71.0}      & 84.5                        & \multicolumn{1}{c|}{8.3}       & \multicolumn{1}{c|}{17.2}      & \multicolumn{1}{c|}{33.1}      & 39.6                       & \multicolumn{1}{c|}{9.8}       & \multicolumn{1}{c|}{21.2}      & \multicolumn{1}{c|}{44.0}   & 54.6                        \\ \hline
DPnet  \cite{tang2023collaborative}                            & \multicolumn{1}{c|}{8.4}       & \multicolumn{1}{c|}{20.5}      & \multicolumn{1}{c|}{53.6}      & 69.1                        & \multicolumn{1}{c|}{25.7}      & \multicolumn{1}{c|}{51.8}      & \multicolumn{1}{c|}{94.9}      & 112.3                       & \multicolumn{1}{c|}{8.3}       & \multicolumn{1}{c|}{18.8}      & \multicolumn{1}{c|}{35.6}      & 44.8                       & \multicolumn{1}{c|}{10.3}      & \multicolumn{1}{c|}{22.9}      & \multicolumn{1}{c|}{47.9}   & 58.1                        \\ \hline
GA-MIN \cite{zhong2023geometric}              & \multicolumn{1}{c|}{7.5}       & \multicolumn{1}{c|}{17.2}      & \multicolumn{1}{c|}{41.1}      & 52.3                        & \multicolumn{1}{c|}{18.9}      & \multicolumn{1}{c|}{38.5}      & \multicolumn{1}{c|}{70.9}      & 84.0                        & \multicolumn{1}{c|}{8.5}       & \multicolumn{1}{c|}{18.3}      & \multicolumn{1}{c|}{34.2}      & 39.9                       & \multicolumn{1}{c|}{9.4}       & \multicolumn{1}{c|}{19.9}      & \multicolumn{1}{c|}{42.4}   & 52.2                        \\ \hline
FMS-AM                               & \multicolumn{1}{c|}{\textbf{5.7}} & \multicolumn{1}{c|}{\textbf{12.1}} & \multicolumn{1}{c|}{\textbf{29.6}} & \textbf{37.4}                   & \multicolumn{1}{c|}{\textbf{14.1}} & \multicolumn{1}{c|}{\textbf{22.8}} & \multicolumn{1}{c|}{\textbf{53.1}} & \textbf{67.5}                   & \multicolumn{1}{c|}{\textbf{6.2}} & \multicolumn{1}{c|}{\textbf{12.5}} & \multicolumn{1}{c|}{\textbf{20.6}} & \textbf{24.5}                  & \multicolumn{1}{c|}{\textbf{6.9}} & \multicolumn{1}{c|}{\textbf{13.4}} & \multicolumn{1}{c|}{\textbf{27.0}}       & \textbf{34.8}                   \\ \hline

\end{tabular}
}

\end{table*}

\begin{table*}[htbp]
\caption{Comparisons between the proposed method and state-of-the-art methods in terms of MPJPE for long-term prediction on 5 action categories of the H3.6M dataset. We also report average performance. The best results are highlighted in bold.}
\label{tab:H3.6M_long_term}
\resizebox{\textwidth}{!}{%
\begin{tabular}{|c|cc|cc|cc|cc|cc|cc|}
\hline
\multirow{2}{*}{Model} & \multicolumn{2}{c|}{walking}               & \multicolumn{2}{c|}{eating}                & \multicolumn{2}{c|}{smoking}               & \multicolumn{2}{c|}{discussion}            & \multicolumn{2}{c|}{directions}      & \multicolumn{2}{c|}{Average}        \\ \cline{2-13} 
                       & \multicolumn{1}{c|}{560}       & 1000      & \multicolumn{1}{c|}{560}       & 1000      & \multicolumn{1}{c|}{560}       & 1000      & \multicolumn{1}{c|}{560}       & 1000      & \multicolumn{1}{c|}{560}    & 1000   & \multicolumn{1}{c|}{560}   & 1000   \\ \hline
Residual Sup \cite{martinez2017human}          & \multicolumn{1}{c|}{81.73}     & 100.68    & \multicolumn{1}{c|}{79.87}     & 100.20    & \multicolumn{1}{c|}{94.83}     & 137.44    & \multicolumn{1}{c|}{121.30}    & 161.70    & \multicolumn{1}{c|}{110.05} & 152.48 & \multicolumn{1}{c|}{97.56} & 130.50 \\ \hline
DMGNN   \cite{li2020dynamic}               & \multicolumn{1}{c|}{73.36}     & 95.82     & \multicolumn{1}{c|}{58.11}     & 86.66     & \multicolumn{1}{c|}{50.85}     & 72.15     & \multicolumn{1}{c|}{81.90}     & 138.32    & \multicolumn{1}{c|}{110.06} & 115.75 & \multicolumn{1}{c|}{74.85} & 101.74 \\ \hline
Traj-GCN \cite{mao2019learning}              & \multicolumn{1}{c|}{54.05}     & 59.75     & \multicolumn{1}{c|}{53.39}     & 77.75     & \multicolumn{1}{c|}{50.74}     & 72.62     & \multicolumn{1}{c|}{91.61}     & 121.53    & \multicolumn{1}{c|}{71.01}  & 101.79 & \multicolumn{1}{c|}{64.16} & 86.69  \\ \hline
MSR-GCN \cite{dang2021msr}               & \multicolumn{1}{c|}{52.72}     & 63.04     & \multicolumn{1}{c|}{52.54}     & 77.11     & \multicolumn{1}{c|}{49.45}     & 71.64     & \multicolumn{1}{c|}{88.59}     & 117.59    & \multicolumn{1}{c|}{71.18}  & 100.59 & \multicolumn{1}{c|}{62.89} & 86.00  \\ \hline
S-DGCN \cite{ma2022progressively}     & \multicolumn{1}{c|}{47.9}      & 55.6      & \multicolumn{1}{c|}{51.2}      & 77.1      & \multicolumn{1}{c|}{45.3}      & 70.3      & \multicolumn{1}{c|}{85.3}      & 110.5     & \multicolumn{1}{c|}{72.5}   & 110.3  & \multicolumn{1}{c|}{60.44} & 84.76  \\ \hline
PK-GCN  \cite{sun2022overlooked}               & \multicolumn{1}{c|}{42.5}      & 47.0      & \multicolumn{1}{c|}{57.9}      & 69.7      & \multicolumn{1}{c|}{33.3}      & 60.2      & \multicolumn{1}{c|}{75.8}      & 112.0     & \multicolumn{1}{c|}{74.7}   & 101.9  & \multicolumn{1}{c|}{56.84} & 78.16  \\ \hline
DANet \cite{cao2022dual}      & \multicolumn{1}{c|}{46.7}      & 55.6      & \multicolumn{1}{c|}{48.0}      & 73.6      & \multicolumn{1}{c|}{44.1}      & 68.0      & \multicolumn{1}{c|}{71.7}      & 108.2     & \multicolumn{1}{c|}{72.9}   & 106.0  & \multicolumn{1}{c|}{56.68} & 82.28  \\ \hline
DPnet \cite{tang2023collaborative}                 & \multicolumn{1}{c|}{40.5}      & 48.6      & \multicolumn{1}{c|}{56.5}      & 69.6      & \multicolumn{1}{c|}{32.8}      & 59.9      & \multicolumn{1}{c|}{66.3}      & 96.7      & \multicolumn{1}{c|}{80.2}   & 103.5  & \multicolumn{1}{c|}{55.26} & 75.66  \\ \hline
GA-MIN \cite{zhong2023geometric}  & \multicolumn{1}{c|}{35.5}      & 42.8      & \multicolumn{1}{c|}{47.3}      & 65.2      & \multicolumn{1}{c|}{30.6}      & 46.5      & \multicolumn{1}{c|}{60.3}      & 106.3     & \multicolumn{1}{c|}{68.1}   & 100.0  & \multicolumn{1}{c|}{48.36} & 72.16  \\ \hline
FMS-AM                   & \multicolumn{1}{c|}{\textbf{23.4}} & \textbf{31.6} & \multicolumn{1}{c|}{\textbf{28.7}} & \textbf{49.3} & \multicolumn{1}{c|}{\textbf{17.8}} & \textbf{38.2} & \multicolumn{1}{c|}{\textbf{39.8}} & \textbf{79.2} & \multicolumn{1}{c|}{\textbf{46.5}}       & \textbf{74.8}       & \multicolumn{1}{c|}{\textbf{31.04}}      &  \textbf{54.62}      \\ \hline
\end{tabular}
}
\end{table*}

Similar, consistent performance of the FMS-AM framework is observed on the CMU-Mocap dataset for both short-term and long-term prediction tasks which are presented in  Tabs. \ref{tab:CMU_short_term} and \ref{tab:CMU_long_term}, respectively. Specifically, the previous state-of-the-art method, GA-MIN \cite{zhong2023geometric}, struggles in short-term prediction of action classes such as jumping and soccer where we observe significant errors $>$ 90\% and 49\%, respectively at the 400 ms setting. Furthermore, for long-term predictions at the 1000 ms setting GA-MIN's average error is $\sim$85\%. In contrast, we observe a significant reduction in prediction error across both short-term and long-term prediction settings in our FMS-AM model where on average it lowers the short-term prediction error by 9.3\% at the 320ms setting, and the long-term prediction at the 1000ms setting is reduced by 32 \%. These results clearly indicate the utility of the proposed innovations.

\begin{figure*}
    \centering
    \includegraphics[width=\linewidth]{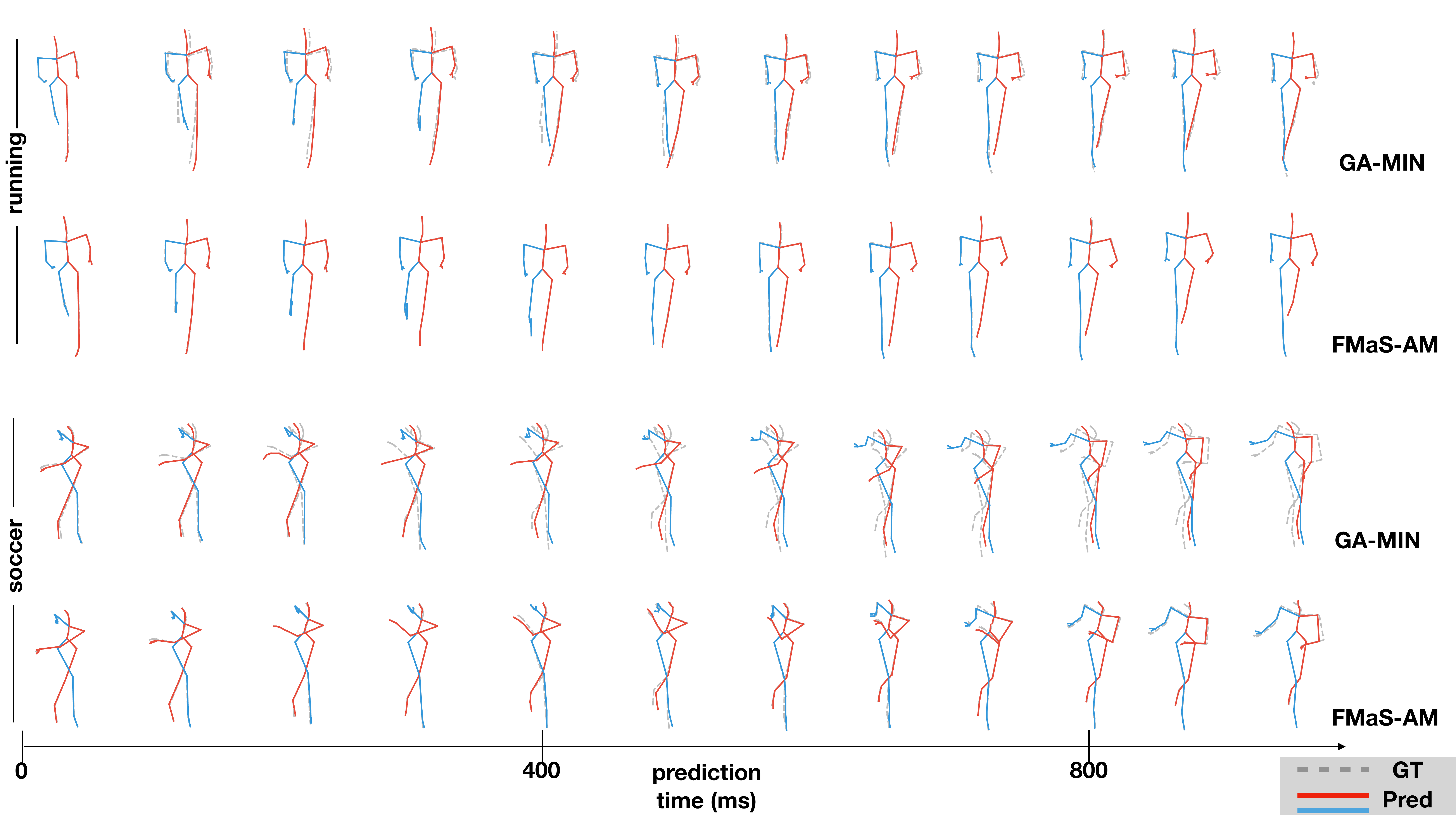}
    \caption{Qualitative Results: A comparison between the existing state-of-the-art model, GA-MIN \cite{zhong2023geometric}, and the proposed FMS-AM method for predicting long-term motion patterns for two action classes in the CMU-Mocap dataset.}
    \label{fig:qualitative}
\end{figure*}

In addition to quantitative comparisons with existing state-of-the-art methods in Fig. \ref{fig:qualitative} we present a comparison between GA-MIN \cite{zhong2023geometric} and the proposed FMS-AM method for predicting long-term motion patterns in the CMU-Mocap dataset for the running and soccer action classes. When comparing the long-term predictions of the two models, in particular beyond 400ms, we observe that GA-MIN struggles to generate accurate estimations of the future motion while the proposed method achieves significant robustness. Furthermore, we observe that the predictions generated by the proposed FMS-AM for complex action classes such as soccer, where there are unique subject-specific and task-specific motions, are far superior to GA-MIN. Please refer to supplementary material for additional qualitative results.

\begin{table*}[htbp]
\caption{Comparisons between the proposed method and state-of-the-art methods in terms of MPJPE for short-term predictions on 8 action categories of the CMU-Mocap dataset. We also report average performance across the action classes. The best results are highlighted in bold.}
\label{tab:CMU_short_term}
\resizebox{\textwidth}{!}{%
\begin{tabular}{|c|cccc|cccc|cccc|cccc|}
\hline
\multirow{2}{*}{Model}             & \multicolumn{4}{c|}{basketball}                                                                                               & \multicolumn{4}{c|}{basketball signal}                                                                                        & \multicolumn{4}{c|}{directing traffic}                                                                                        & \multicolumn{4}{c|}{jumping}                                                                                               \\ \cline{2-17} 
                                   & \multicolumn{1}{c|}{80}        & \multicolumn{1}{c|}{160}       & \multicolumn{1}{c|}{320}       & 400                        & \multicolumn{1}{c|}{80}        & \multicolumn{1}{c|}{160}       & \multicolumn{1}{c|}{320}       & 400                        & \multicolumn{1}{c|}{80}        & \multicolumn{1}{c|}{160}       & \multicolumn{1}{c|}{320}       & 400                        & \multicolumn{1}{c|}{80}        & \multicolumn{1}{c|}{160}       & \multicolumn{1}{c|}{320}   & 400                         \\ \hline
\multicolumn{1}{|l|}{Residual Sup \cite{martinez2017human}} & \multicolumn{1}{l|}{15.45}     & \multicolumn{1}{l|}{26.88}     & \multicolumn{1}{l|}{43.51}     & \multicolumn{1}{l|}{49.23} & \multicolumn{1}{l|}{20.17}     & \multicolumn{1}{l|}{32.98}     & \multicolumn{1}{l|}{42.75}     & \multicolumn{1}{l|}{44.65} & \multicolumn{1}{l|}{20.52}     & \multicolumn{1}{l|}{40.58}     & \multicolumn{1}{l|}{75.38}     & \multicolumn{1}{l|}{90.36} & \multicolumn{1}{l|}{26.85}     & \multicolumn{1}{l|}{48.07}     & \multicolumn{1}{l|}{93.50} & \multicolumn{1}{l|}{108.90} \\ \hline
DMGNN  \cite{li2020dynamic}                            & \multicolumn{1}{c|}{15.57}     & \multicolumn{1}{c|}{28.72}     & \multicolumn{1}{c|}{59.01}     & 73.05                      & \multicolumn{1}{c|}{5.03}      & \multicolumn{1}{c|}{9.28}      & \multicolumn{1}{c|}{20.21}     & 26.23                      & \multicolumn{1}{c|}{10.21}     & \multicolumn{1}{c|}{20.90}     & \multicolumn{1}{c|}{41.55}     & 52.28                      & \multicolumn{1}{c|}{31.97}     & \multicolumn{1}{c|}{54.32}     & \multicolumn{1}{c|}{96.66} & 119.92                      \\ \hline
Traj-GCN   \cite{mao2019learning}                        & \multicolumn{1}{c|}{11.68}     & \multicolumn{1}{c|}{21.26}     & \multicolumn{1}{c|}{40.99}     & 50.78                      & \multicolumn{1}{c|}{3.33}      & \multicolumn{1}{c|}{6.25}      & \multicolumn{1}{c|}{13.58}     & 17.98                      & \multicolumn{1}{c|}{6.92}      & \multicolumn{1}{c|}{13.69}     & \multicolumn{1}{c|}{30.30}     & 39.97                      & \multicolumn{1}{c|}{17.18}     & \multicolumn{1}{c|}{32.37}     & \multicolumn{1}{c|}{60.12} & 72.55                       \\ \hline
MSR-GCN \cite{dang2021msr}                           & \multicolumn{1}{c|}{10.28}     & \multicolumn{1}{c|}{18.94}     & \multicolumn{1}{c|}{37.68}     & 47.03                      & \multicolumn{1}{c|}{3.03}      & \multicolumn{1}{c|}{5.68}      & \multicolumn{1}{c|}{12.35}     & 16.26                      & \multicolumn{1}{c|}{5.92}      & \multicolumn{1}{c|}{12.09}     & \multicolumn{1}{c|}{28.36}     & 38.04                      & \multicolumn{1}{c|}{14.99}     & \multicolumn{1}{c|}{28.66}     & \multicolumn{1}{c|}{55.86} & 62.93                       \\ \hline
DPnet \cite{tang2023collaborative}                             & \multicolumn{1}{c|}{10.7}      & \multicolumn{1}{c|}{17.8}      & \multicolumn{1}{c|}{38.4}      & 49.5                       & \multicolumn{1}{c|}{2.6}       & \multicolumn{1}{c|}{4.4}       & \multicolumn{1}{c|}{10.0}      & 13.4                       & \multicolumn{1}{c|}{5.9}       & \multicolumn{1}{c|}{11.8}      & \multicolumn{1}{c|}{26.6}      & 33.5                       & \multicolumn{1}{c|}{12.4}      & \multicolumn{1}{c|}{28.3}      & \multicolumn{1}{c|}{70.2}  & 89.2                        \\ \hline
GA-MIN \cite{zhong2023geometric}              & \multicolumn{1}{c|}{10.3}      & \multicolumn{1}{c|}{19.8}      & \multicolumn{1}{c|}{40.3}      & 51.8                       & \multicolumn{1}{c|}{2.5}       & \multicolumn{1}{c|}{4.6}       & \multicolumn{1}{c|}{10.5}      & 15.3                       & \multicolumn{1}{c|}{\textbf{5.7}}       & \multicolumn{1}{c|}{10.8}      & \multicolumn{1}{c|}{27.2}      & 33.4                       & \multicolumn{1}{c|}{14.2}      & \multicolumn{1}{c|}{28.2}      & \multicolumn{1}{c|}{71.8}  & 91.1                        \\ \hline
FMS-AM                               & \multicolumn{1}{c|}{\textbf{6.2}} & \multicolumn{1}{c|}{\textbf{14.6}} & \multicolumn{1}{c|}{\textbf{26.2}} & \textbf{34.1}                  & \multicolumn{1}{c|}{\textbf{2.4}} & \multicolumn{1}{c|}{\textbf{3.2}} & \multicolumn{1}{c|}{\textbf{6.8}} & \textbf{10.6}                  & \multicolumn{1}{c|}{5.8} & \multicolumn{1}{c|}{\textbf{8.4}} & \multicolumn{1}{c|}{\textbf{19.8}} & \textbf{26.4}                  & \multicolumn{1}{c|}{\textbf{9.3}} & \multicolumn{1}{c|}{\textbf{16.3}} & \multicolumn{1}{c|}{\textbf{56.2}}      & \textbf{79.2}                   \\ \hline

\hline
\hline
\multirow{2}{*}{Model} & \multicolumn{4}{c|}{running}                                                                                 & \multicolumn{4}{c|}{soccer}                                                                                  & \multicolumn{4}{c|}{walking}                                                                                 & \multicolumn{4}{c|}{washwindow}                                                                          \\ \cline{2-17} 
                       & \multicolumn{1}{c|}{80}        & \multicolumn{1}{c|}{160}       & \multicolumn{1}{c|}{320}       & 400       & \multicolumn{1}{c|}{80}        & \multicolumn{1}{c|}{160}       & \multicolumn{1}{c|}{320}       & 400       & \multicolumn{1}{c|}{80}        & \multicolumn{1}{c|}{160}       & \multicolumn{1}{c|}{320}       & 400       & \multicolumn{1}{c|}{80}        & \multicolumn{1}{c|}{160}       & \multicolumn{1}{c|}{320}   & 400       \\ \hline
Residual Sup \cite{martinez2017human}          & \multicolumn{1}{c|}{25.76}     & \multicolumn{1}{c|}{48.91}     & \multicolumn{1}{c|}{88.19}     & 100.80    & \multicolumn{1}{c|}{17.75}     & \multicolumn{1}{c|}{31.30}     & \multicolumn{1}{c|}{52.55}     & 61.40     & \multicolumn{1}{c|}{44.35}     & \multicolumn{1}{c|}{76.66}     & \multicolumn{1}{c|}{216.83}    & 151.43    & \multicolumn{1}{c|}{22.84}     & \multicolumn{1}{c|}{44.71}     & \multicolumn{1}{c|}{86.78} & 104.68    \\ \hline
DMGNN  \cite{li2020dynamic}                & \multicolumn{1}{c|}{17.42}     & \multicolumn{1}{c|}{26.82}     & \multicolumn{1}{c|}{38.27}     & 40.08     & \multicolumn{1}{c|}{14.86}     & \multicolumn{1}{c|}{25.29}     & \multicolumn{1}{c|}{52.21}     & 26.23     & \multicolumn{1}{c|}{9.57}      & \multicolumn{1}{c|}{15.53}     & \multicolumn{1}{c|}{26.03}     & 30.37     & \multicolumn{1}{c|}{7.93}      & \multicolumn{1}{c|}{14.68}     & \multicolumn{1}{c|}{33.34} & 44.24     \\ \hline
Traj-GCN \cite{mao2019learning}              & \multicolumn{1}{c|}{14.53}     & \multicolumn{1}{c|}{24.20}     & \multicolumn{1}{c|}{37.44}     & 41.10     & \multicolumn{1}{c|}{3.33}      & \multicolumn{1}{c|}{6.25}      & \multicolumn{1}{c|}{13.58}     & 17.98     & \multicolumn{1}{c|}{6.62}      & \multicolumn{1}{c|}{10.74}     & \multicolumn{1}{c|}{17.40}     & 20.35     & \multicolumn{1}{c|}{5.96}      & \multicolumn{1}{c|}{11.62}     & \multicolumn{1}{c|}{24.77} & 31.63     \\ \hline
MSR-GCN \cite{dang2021msr}               & \multicolumn{1}{c|}{12.84}     & \multicolumn{1}{c|}{20.42}     & \multicolumn{1}{c|}{30.58}     & 34.42     & \multicolumn{1}{c|}{3.03}      & \multicolumn{1}{c|}{5.68}      & \multicolumn{1}{c|}{12.35}     & 16.26     & \multicolumn{1}{c|}{6.31}      & \multicolumn{1}{c|}{10.30}     & \multicolumn{1}{c|}{17.64}     & 21.12     & \multicolumn{1}{c|}{5.49}      & \multicolumn{1}{c|}{11.07}     & \multicolumn{1}{c|}{25.05} & 32.51     \\ \hline
DPnet  \cite{tang2023collaborative}                & \multicolumn{1}{c|}{16.7}      & \multicolumn{1}{c|}{18.4}      & \multicolumn{1}{c|}{19.6}      & 25.1      & \multicolumn{1}{c|}{9.0}       & \multicolumn{1}{c|}{17.1}      & \multicolumn{1}{c|}{35.8}      & 48.7      & \multicolumn{1}{c|}{5.8}       & \multicolumn{1}{c|}{9.0}       & \multicolumn{1}{c|}{17.2}      & 21.4      & \multicolumn{1}{c|}{4.5}       & \multicolumn{1}{c|}{\textbf{9.8}}       & \multicolumn{1}{c|}{27.3}  & 36.7      \\ \hline
GA-MIN \cite{zhong2023geometric}  & \multicolumn{1}{c|}{17.5}      & \multicolumn{1}{c|}{22.3}      & \multicolumn{1}{c|}{22.1}      & 26.1      & \multicolumn{1}{c|}{9.8}       & \multicolumn{1}{c|}{18.3}      & \multicolumn{1}{c|}{39.0}      & 49.4      & \multicolumn{1}{c|}{5.2}       & \multicolumn{1}{c|}{8.9}       & \multicolumn{1}{c|}{16.2}      & 18.2      & \multicolumn{1}{c|}{4.5}       & \multicolumn{1}{c|}{9.9}       & \multicolumn{1}{c|}{27.8}  & 35.2      \\ \hline
FMS-AM                   & \multicolumn{1}{c|}{\textbf{10.1}} & \multicolumn{1}{c|}{\textbf{12.6}} & \multicolumn{1}{c|}{\textbf{15.3}} & \textbf{18.8} & \multicolumn{1}{c|}{\textbf{6.7}} & \multicolumn{1}{c|}{\textbf{11.2}} & \multicolumn{1}{c|}{\textbf{26.8}} & \textbf{31.4} & \multicolumn{1}{c|}{\textbf{3.9}} & \multicolumn{1}{c|}{\textbf{6.1}} & \multicolumn{1}{c|}{\textbf{11.8}} & \textbf{14.9} & \multicolumn{1}{c|}{\textbf{4.4}} & \multicolumn{1}{c|}{10.0} & \multicolumn{1}{c|}{\textbf{18.1}}      & \textbf{21.7} \\ \hline
\end{tabular}
}
\end{table*}

\begin{table}[htbp]
\caption{ Comparisons between the proposed method and state-of-the-art methods in terms of average MPJPE for short-term and long-term predictions of the CMU-Mocap dataset. The best results are highlighted in bold. Note that `-' represents the unavailability of baseline evaluations in that setting.}
\label{tab:CMU_long_term}
\begin{tabular}{|c|cccccc|}
\hline
\multirow{2}{*}{Model} & \multicolumn{6}{c|}{Average}                                                                                                                      \\ \cline{2-7} 
                       & \multicolumn{1}{c|}{80}   & \multicolumn{1}{c|}{160}  & \multicolumn{1}{c|}{320}  & \multicolumn{1}{c|}{400}  & \multicolumn{1}{c|}{560}  & 1000  \\ \hline
DMGNN  \cite{li2020dynamic}                & \multicolumn{1}{c|}{13.6} & \multicolumn{1}{c|}{24.1} & \multicolumn{1}{c|}{47.0} & \multicolumn{1}{c|}{58.8} & \multicolumn{1}{c|}{77.4} & 112.6 \\ \hline
Traj-GCN   \cite{mao2019learning}            & \multicolumn{1}{c|}{11.2} & \multicolumn{1}{c|}{19.1} & \multicolumn{1}{c|}{36.3} & \multicolumn{1}{c|}{-}    & \multicolumn{1}{c|}{45.8} & 95.7  \\ \hline
S-DGCN \cite{ma2022progressively}     & \multicolumn{1}{c|}{7.6}  & \multicolumn{1}{c|}{14.3} & \multicolumn{1}{c|}{29.0} & \multicolumn{1}{c|}{36.6} & \multicolumn{1}{c|}{50.9} & 80.1  \\ \hline
PK-GCN  \cite{sun2022overlooked}               & \multicolumn{1}{c|}{9.4}  & \multicolumn{1}{c|}{17.1} & \multicolumn{1}{c|}{32.8} & \multicolumn{1}{c|}{40.3} & \multicolumn{1}{c|}{52.2} & 79.3  \\ \hline
DPnet \cite{tang2023collaborative}                 & \multicolumn{1}{c|}{8.4}  & \multicolumn{1}{c|}{14.5} & \multicolumn{1}{c|}{30.6} & \multicolumn{1}{c|}{39.7} & \multicolumn{1}{c|}{-}    & 91.3  \\ \hline
GA-MIN \cite{zhong2023geometric}  & \multicolumn{1}{c|}{8.7}  & \multicolumn{1}{c|}{15.4} & \multicolumn{1}{c|}{31.9} & \multicolumn{1}{c|}{-}    & \multicolumn{1}{c|}{40.1} & 84.5  \\
\hline
FMS-AM                   & \multicolumn{1}{c|}{\textbf{6.1}}     & \multicolumn{1}{c|}{\textbf{10.3}}     & \multicolumn{1}{c|}{\textbf{22.6}}     & \multicolumn{1}{c|}{\textbf{29.6}}     & \multicolumn{1}{c|}{\textbf{32.4}}     &   \textbf{52.5}    \\ 
\hline
\end{tabular}
\end{table}

\subsection{Ablation Evaluations}\label{Sec:ablations}

We conducted a series of ablation studies to systematically analyse the impact of the individual innovations that our FMS-AM framework proposes. Several design choices contribute to the robustness of our model: i) the proposed feature factorisation strategy; ii) the multi-head knowledge retrieval scheme; iii) the proposed auxiliary memory stabilisation losses; and iv) the dynamic mask generation procedure. All of these experiments were conducted on the H3.6M dataset and for testing the ablation models we use the validation set of H3.6M. 

\subsubsection{Effects of Feature Factorisation}

To study the effect of the proposed feature factorisation strategy we generated four ablation variants of the proposed FMS-AM model: i) S-AM- w/o [F,Mh]: a model with an auxiliary memory and proposed stabilisation losses, but without feature factorisation and the multi-head knowledge retrieval scheme, ii) MhS-AM - 2F v1, iii) MhS-AM - 2F v2, and iv) MhS-AM - 2F v3 are models with an auxiliary memory, the proposed multi-head knowledge retrieval scheme and proposed stabilisation losses. They also possess the ability to factorise features. However, they only factorise the features into two factors. Please refer to Tab. \ref{tab:factorisation_ablation} for details regarding the two factors that each of these models consider.

\begin{table*}[htbp]
\caption{Effect of the Proposed Feature Factorisation Scheme}
\label{tab:factorisation_ablation}
\resizebox{\textwidth}{!}{%
\begin{tabular}{|c|c|ccc|c|cc|cccccc|}
\hline
\multirow{2}{*}{Model} & \multirow{2}{*}{Auxiliary Memory}                       & \multicolumn{3}{c|}{Factorisation}                                                   & \multirow{2}{*}{Multi-Head Access} & \multicolumn{2}{c|}{Losses}                       & \multicolumn{6}{c|}{Average}                                                                                                                                              \\ \cline{3-5} \cline{7-14} 
                       &                                                         & \multicolumn{1}{c|}{Subject}      & \multicolumn{1}{c|}{Task}         & Auxiliary    &                                              & \multicolumn{1}{c|}{Diversity}    & Stabilisation & \multicolumn{1}{c|}{80} & \multicolumn{1}{c|}{160} & \multicolumn{1}{c|}{320} & \multicolumn{1}{c|}{400} & \multicolumn{1}{c|}{560} & \multicolumn{1}{c|}{1000} \\ \hline
S-AM w/o [F, Mh]        & $\checkmark$ & \multicolumn{1}{c|}{}             & \multicolumn{1}{c|}{}             &              &                                              & \multicolumn{1}{c|}{$\checkmark$} & $\checkmark$  & \multicolumn{1}{c|}{9.5}   & \multicolumn{1}{c|}{23.2}    & \multicolumn{1}{c|}{48.8}    & \multicolumn{1}{c|}{58.4}    & \multicolumn{1}{c|}{57.4}    & \multicolumn{1}{c|}{78.7}             \\ \hline
MhS-AM - 2F v1             & $\checkmark$                                            & \multicolumn{1}{c|}{$\checkmark$} & \multicolumn{1}{c|}{}             & $\checkmark$ & $\checkmark$                                 & \multicolumn{1}{c|}{$\checkmark$} & $\checkmark$  & \multicolumn{1}{c|}{7.8}   & \multicolumn{1}{c|}{28.7}    & \multicolumn{1}{c|}{29.2}    & \multicolumn{1}{c|}{36.8}    & \multicolumn{1}{c|}{35.8}    & \multicolumn{1}{c|}{59.1}            \\ \hline
MhS-AM - 2F v2             & $\checkmark$                                            & \multicolumn{1}{c|}{}             & \multicolumn{1}{c|}{$\checkmark$} & $\checkmark$ & $\checkmark$                                 & \multicolumn{1}{c|}{$\checkmark$} & $\checkmark$  & \multicolumn{1}{c|}{7.5}   & \multicolumn{1}{c|}{17.3}    & \multicolumn{1}{c|}{28.4}    & \multicolumn{1}{c|}{36.8}    & \multicolumn{1}{c|}{35.7}    & \multicolumn{1}{c|}{58.9}           \\ \hline
MhS-AM - 2F v3             & $\checkmark$                                            & \multicolumn{1}{c|}{$\checkmark$} & \multicolumn{1}{c|}{$\checkmark$}             &  & $\checkmark$                                 & \multicolumn{1}{c|}{$\checkmark$} & $\checkmark$  & \multicolumn{1}{c|}{7.0}   & \multicolumn{1}{c|}{15.4}    & \multicolumn{1}{c|}{27.2}    & \multicolumn{1}{c|}{35.4}    & \multicolumn{1}{c|}{34.2}    & \multicolumn{1}{c|}{55.6}           \\ \hline
FMS-AM                & $\checkmark$                                            & \multicolumn{1}{c|}{$\checkmark$} & \multicolumn{1}{c|}{$\checkmark$} & $\checkmark$ & $\checkmark$                                 & \multicolumn{1}{c|}{$\checkmark$} & $\checkmark$  & \multicolumn{1}{c|}{6.8}   & \multicolumn{1}{c|}{13.5}    & \multicolumn{1}{c|}{26.8}    & \multicolumn{1}{c|}{33.8}    & \multicolumn{1}{c|}{32.6}    & \multicolumn{1}{c|}{53.9}           \\ \hline
\end{tabular}}
\end{table*}

Comparison results are shown in Tab. \ref{tab:factorisation_ablation}. We observe that the feature factorisation procedure is an integral part of the proposed framework. In particular, we observe that the auxiliary memory struggles to reach significant robustness levels without the feature factorisation scheme. With the introduction of two factorised levels, we observe higher levels of robustness where subject and task-specific attributes seem to make a significant impact (see the rows corresponding to MhS-AM - 2F v1, MhS-AM - 2F v2, and MhS-AM - 2F v3). Furthermore, we observe that the proposed stabilisation losses are capable of better aiding the learning process when the extracted feature representation is factorised. These experiments demonstrate the utility of our feature factorisation procedure.

\begin{figure}[htbp]
    \centering
     \begin{subfigure}[b]{0.45\textwidth}
         \centering
         \includegraphics[width=\textwidth]{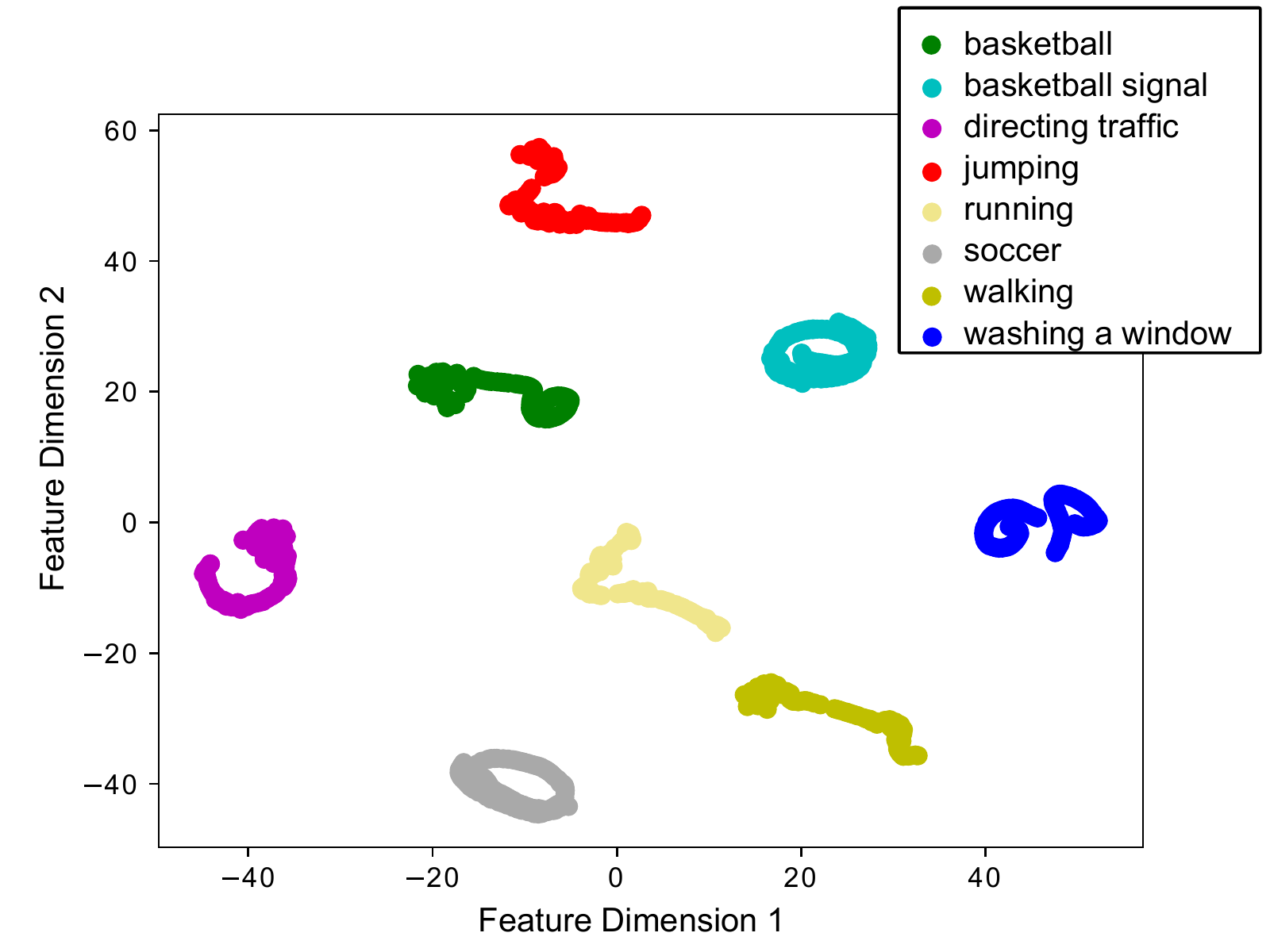}
         \caption{Factorised Task Embeddings}
     \end{subfigure}
     \hfill
     \begin{subfigure}[b]{0.45\textwidth}
         \centering
         \includegraphics[width=\textwidth]{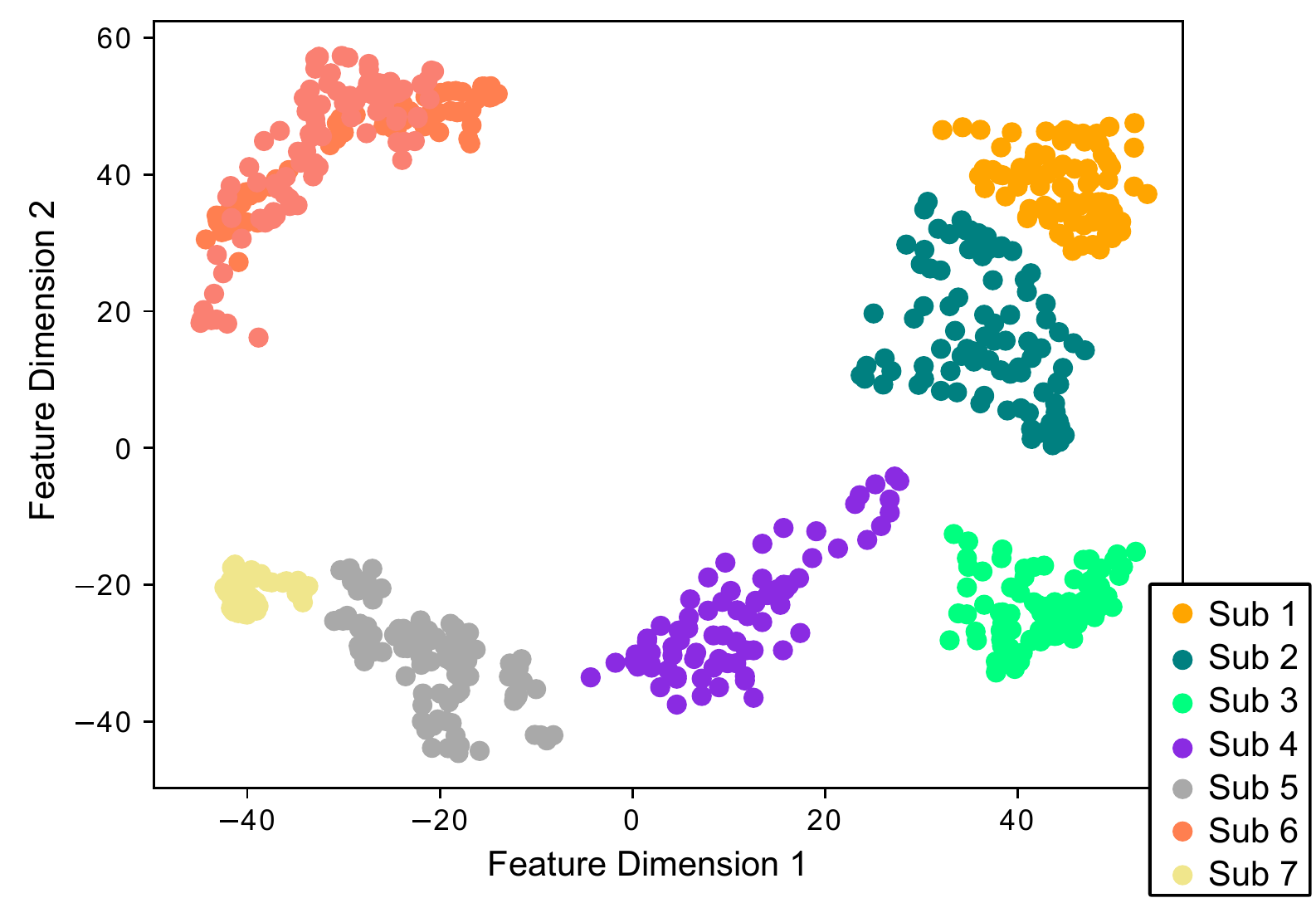}
         \caption{Factorised Subject Embeddings}
     \end{subfigure}
    \caption{2D projections obtained using t-SNE of the factorised subject-specific and task-specific features}
    \label{fig:tsne}
\end{figure}

In addition to qualitative results, in Fig. \ref{fig:tsne} we visualise the 2D projections of the factorised subject-specific and task-specific embeddings. In this experiment we utilise 200 randomly chosen samples from the CMU-Mocap dataset and use t-SNE for 2D projection of the factorised features. From Fig. \ref{fig:tsne} it is clear that the proposed factorisation strategy has been able to clearly separate the feature space into unique subject-specific and task-specific sub-spaces.

\subsubsection{Effects of Multi-Head Knowledge Retrieval and Auxiliary Memory Stabilisation Losses}

The effectiveness of the proposed multi-head knowledge retrieval scheme and the auxiliary memory stabilisation losses are evaluated in this experiment. We generated five ablation variants of the proposed FMS-AM model: i) F w/o [AM, Mh, S]: a model using the proposed feature factorisation, but without the auxiliary memory, the proposed stabilisation losses and multi-head knowledge retrieval scheme; ii) F-AM w/o [Mh, S]: this model extends the model in i) adding an auxiliary memory; iii) FMh-AM w/o S: the proposed model without the stabilisation and diversity losses; iv) FMh-AM-S v1: the proposed model with the diversity loss, but without the stabilisation loss; and v) FMh-AM-S v2: the proposed model with the stabilisation loss, but without the diversity loss.

\begin{table*}[htbp]
\caption{Effects of the Proposed Multi Head Knowledge Retrieval Scheme and Auxiliary Memory Stabilisation Losses}
\label{tab:ablation_Ma_and_S}
\resizebox{\textwidth}{!}{%
\begin{tabular}{|c|c|ccc|c|cc|cccccc|}
\hline
\multirow{2}{*}{Model} & \multirow{2}{*}{Auxiliary Memory} & \multicolumn{3}{c|}{Factorisation}                                                   & \multirow{2}{*}{Multi Head Access} & \multicolumn{2}{c|}{Losses}                       & \multicolumn{6}{c|}{Average}                                                                                                                                              \\ \cline{3-5} \cline{7-14} 
                       &                                   & \multicolumn{1}{c|}{Subject}      & \multicolumn{1}{c|}{Task}         & Auxiliary    &                                              & \multicolumn{1}{c|}{Diversity}    & Stabilisation & \multicolumn{1}{c|}{80} & \multicolumn{1}{c|}{160} & \multicolumn{1}{c|}{320} & \multicolumn{1}{c|}{400} & \multicolumn{1}{c|}{560} & \multicolumn{1}{c|}{1000} \\ \hline
F w/o {[}AM, Mh, S{]}  &                                   & \multicolumn{1}{c|}{$\checkmark$} & \multicolumn{1}{c|}{$\checkmark$} & $\checkmark$ &                                              & \multicolumn{1}{c|}{}             &               & \multicolumn{1}{c|}{10.0}   & \multicolumn{1}{c|}{24.1}    & \multicolumn{1}{c|}{48.2}    & \multicolumn{1}{c|}{58.4}    & \multicolumn{1}{c|}{64.3}    & \multicolumn{1}{c|}{85.8}            \\ \hline
F-AM  w/o {[}Mh, S{]}  & $\checkmark$                      & \multicolumn{1}{c|}{$\checkmark$} & \multicolumn{1}{c|}{$\checkmark$} & $\checkmark$ &                                              & \multicolumn{1}{c|}{}             &               & \multicolumn{1}{c|}{9.1}   & \multicolumn{1}{c|}{22.1}    & \multicolumn{1}{c|}{43.8}    & \multicolumn{1}{c|}{47.5}    & \multicolumn{1}{c|}{56.2}    & \multicolumn{1}{c|}{76.5}            \\ \hline
FMh-AM w/o S           & $\checkmark$                      & \multicolumn{1}{c|}{$\checkmark$} & \multicolumn{1}{c|}{$\checkmark$} & $\checkmark$ & $\checkmark$                                 & \multicolumn{1}{c|}{}             &               & \multicolumn{1}{c|}{8.2}   & \multicolumn{1}{c|}{20.2}    & \multicolumn{1}{c|}{35.4}    & \multicolumn{1}{c|}{38.3}    & \multicolumn{1}{c|}{35.4}    & \multicolumn{1}{c|}{60.2}            \\ \hline
FMh-AM - S v1          & $\checkmark$                      & \multicolumn{1}{c|}{$\checkmark$} & \multicolumn{1}{c|}{$\checkmark$} & $\checkmark$ & $\checkmark$                                 & \multicolumn{1}{c|}{$\checkmark$} &               & \multicolumn{1}{c|}{7.4}   & \multicolumn{1}{c|}{15.8}    & \multicolumn{1}{c|}{28.1}    & \multicolumn{1}{c|}{36.4}    & \multicolumn{1}{c|}{34.2}    & \multicolumn{1}{c|}{56.8}             \\ \hline
FMh-AM - S v2          & $\checkmark$                      & \multicolumn{1}{c|}{$\checkmark$} & \multicolumn{1}{c|}{$\checkmark$} & $\checkmark$ & $\checkmark$                                 & \multicolumn{1}{c|}{}             & $\checkmark$  & \multicolumn{1}{c|}{7.3}   & \multicolumn{1}{c|}{15.6}    & \multicolumn{1}{c|}{27.8}    & \multicolumn{1}{c|}{36.1}    & \multicolumn{1}{c|}{34.5}    & \multicolumn{1}{c|}{56.2}             \\ \hline
FMS-AM                & $\checkmark$                      & \multicolumn{1}{c|}{$\checkmark$} & \multicolumn{1}{c|}{$\checkmark$} & $\checkmark$ & $\checkmark$                                 & \multicolumn{1}{c|}{$\checkmark$} & $\checkmark$  & \multicolumn{1}{c|}{6.8}   & \multicolumn{1}{c|}{13.5}    & \multicolumn{1}{c|}{26.8}    & \multicolumn{1}{c|}{33.8}    & \multicolumn{1}{c|}{32.6}    & \multicolumn{1}{c|}{53.9}            \\ \hline
\end{tabular}}
\end{table*}

From the results in Tab. \ref{tab:ablation_Ma_and_S} we can confirm that there is a utility with respect to incorporating both subject-specific and task-specific historical knowledge captured by the auxiliary memory into the prediction framework, as evidenced by the rows in Tab. \ref{tab:ablation_Ma_and_S} corresponding to models F w/o [AM, Mh, S] and F-AM w/o [Mh, S]. Furthermore, the experimental results presented in Tab. \ref{tab:ablation_Ma_and_S} confirm our hypothesis that both multi-head knowledge retrieval and the proposed stabilisation losses contribute to the robustness of our model. We refer the reader to the rows corresponding to F-AM w/o [Mh, S] and FMh-AM w/o S in Tab. \ref{tab:ablation_Ma_and_S}, where we observe a significant increase in the accuracy with the introduction of multi-head knowledge retrieval. This is because by utilising multiple factorised information cues, the read operation of our auxiliary memory can retrieve multiple items of salient information to aid prediction. Furthermore, rows corresponding to FMh-AM w/o S and FMS-AM confirm that stabilisation losses are integral parts of our framework. Moreover, we observe that both diversity and stabilisation losses are pivotal for the convergence of the knowledge captured within the auxiliary memory. Therefore, using this experiment we can confirm the necessity of both the multi-head knowledge retrieval scheme and the auxiliary memory stabilitation losses within our framework. 

\subsubsection{Effects of Dynamic Mask Generation Procedure}

To better establish the contributions of the proposed dynamic mask generation procedure we conducted an additional ablation experiment. Note that in the previous two ablation experiments in situations where factorisation of features is leveraged, we utilise a dynamic masking strategy. However, in this experiment, we utilise both static masks (shown in Fig. \ref{fig:fixed_masks}) and the dynamic masks generated using the proposed mask generation procedure. Specifically, three ablation variants of the proposed model were generated: i) FS-AM w/o [Mh, DM]: the proposed model without multi-head retrieval and dynamic masking; ii) FMS-AM w/o DM: the model of i) with multi-head retrieval but without dynamic masking; and iii) [FS-AM, DM] w/o Mh: the model of i) with dynamic masking but without multi-head retrieval.

\begin{table*}[htbp]
\caption{Effects of the Proposed Dynamic Mask Generation Procedure}
\label{tab:ablation_dynamic_masks}
\resizebox{\textwidth}{!}{%
\begin{tabular}{|c|c|ccc|c|cc|c|cccccc|}
\hline
\multirow{2}{*}{Model} & \multirow{2}{*}{Auxiliary Memory} & \multicolumn{3}{c|}{Factorisation}                                                   & \multirow{2}{*}{Multi-Head Access} & \multicolumn{2}{c|}{Losses}                       & \multirow{2}{*}{Dynamic Masking} & \multicolumn{6}{c|}{Average}                                                                                                                                              \\ \cline{3-5} \cline{7-8} \cline{10-15} 
                       &                                   & \multicolumn{1}{c|}{Subject}      & \multicolumn{1}{c|}{Task}         & Auxiliary    &                                              & \multicolumn{1}{c|}{Diversity}    & Stabilisation &                                  & \multicolumn{1}{c|}{80} & \multicolumn{1}{c|}{160} & \multicolumn{1}{c|}{320} & \multicolumn{1}{c|}{400} & \multicolumn{1}{c|}{560} & \multicolumn{1}{c|}{1000} \\ \hline
FS-AM w/o {[}Mh, DM{]} & $\checkmark$                      & \multicolumn{1}{c|}{$\checkmark$} & \multicolumn{1}{c|}{$\checkmark$} & $\checkmark$ &                                              & \multicolumn{1}{c|}{$\checkmark$} & $\checkmark$  &                                  & \multicolumn{1}{c|}{8.9}   & \multicolumn{1}{c|}{21.8}    & \multicolumn{1}{c|}{40.4}    & \multicolumn{1}{c|}{47.2}    & \multicolumn{1}{c|}{54.6}    & \multicolumn{1}{c|}{75.5}             \\ \hline
FMS-AM w/o DM         & $\checkmark$                      & \multicolumn{1}{c|}{$\checkmark$} & \multicolumn{1}{c|}{$\checkmark$} & $\checkmark$ & $\checkmark$                                 & \multicolumn{1}{c|}{$\checkmark$} & $\checkmark$         &                          & \multicolumn{1}{c|}{8.5}   & \multicolumn{1}{c|}{18.2}    & \multicolumn{1}{c|}{34.4}    & \multicolumn{1}{c|}{37.6}    & \multicolumn{1}{c|}{38.3}    & \multicolumn{1}{c|}{64.5}             \\ \hline
{[}FS-AM, DM{]} w/o Mh & $\checkmark$                      & \multicolumn{1}{c|}{$\checkmark$} & \multicolumn{1}{c|}{$\checkmark$} & $\checkmark$                                              & \multicolumn{1}{c|}{$\checkmark$} & $\checkmark$  & $\checkmark$        &             & \multicolumn{1}{c|}{8.8}   & \multicolumn{1}{c|}{20.4}    & \multicolumn{1}{c|}{38.9}    & \multicolumn{1}{c|}{45.4}    & \multicolumn{1}{c|}{50.5}    & \multicolumn{1}{c|}{73.1}             \\ \hline
FMS-AM                & $\checkmark$                      & \multicolumn{1}{c|}{$\checkmark$} & \multicolumn{1}{c|}{$\checkmark$} & $\checkmark$ & $\checkmark$                                 & \multicolumn{1}{c|}{$\checkmark$} & $\checkmark$  & $\checkmark$                     & \multicolumn{1}{c|}{6.8}   & \multicolumn{1}{c|}{13.5}    & \multicolumn{1}{c|}{26.8}    & \multicolumn{1}{c|}{33.8}    & \multicolumn{1}{c|}{32.6}    & \multicolumn{1}{c|}{53.9}             \\ \hline
\end{tabular}}
\end{table*}

The results presented in Tab. \ref{tab:ablation_dynamic_masks} confirm the need for dynamic masking. In particular, the ablation model without both the multi-head retrieval mechanism and the dynamic masking procedure (see the row corresponding to FS-AM w/o [Mh, DM] in Tab. \ref{tab:ablation_dynamic_masks}) significantly struggles to achieve good prediction results. Furthermore, comparing rows corresponding to FMS-AM w/o DM and [FS-AM, DM] w/o Mh models in Tab. \ref{tab:ablation_dynamic_masks} we can confirm that multi-head retrieval and dynamic masking complement each other. This is because depending on the context of the input partial activations and dynamic changes of the mask are possible, and by leveraging this dynamic mask the multi-head retrieval scheme can retrieve more informative content from the auxiliary memory. These observations strongly validate the importance of the proposed dynamic mask generation procedure. 

\subsection{Time Complexity}\label{sec:time_complexity}
 Our FMS-AM method contains 30.82M trainable parameters which is not a substantial increase of trainable parameters when compared with the state-of-the-art MGCN model \cite{zhou2021learning} which has 26.51M trainable parameters when considering the significant performance increase that our FMS-AM method achieves. FMS-AM generates predictions (each of which has a prediction length of 1000 ms) for 100 input pose sequences in 5.85675 sec using a single NVIDIA A100 GPU. Note that this includes time taken for both feature extraction and the generation of model predictions.  

\section{Conclusion}
This paper presented a Factorised Multi-head retrieval and Stabilisation based Auxiliary Memory (FMS-AM) powered framework for the accurate prediction of future human motions. We demonstrated that dynamic factorisation of the subject-specific and task-specific features plays a pivotal role in the effectiveness of the proposed architecture. Furthermore, our innovative multi-head knowledge retrieval scheme that leverages these factorised embeddings to generate multiple query operations is an integral part of our framework. The introduced loss functions guarantee that the knowledge captured within the auxiliary memory is not influenced by data imbalances or the diversity of the input data distributions. Extensive experiments were conducted on two public benchmarks: Human3.6M and CMU-Mocap, which demonstrated the ability of the proposed framework to outperform the current state-of-the-art algorithms by significant margins. 

We observe two limitations of FMS-AM and suggest the following future research directions: (i) This study has only investigated the use of single auxiliary memory that stores  subject-specific, task-specific, and auxiliary features together. The dynamic masking strategy is leveraged to query multiple representations out of the memory. In future works, we will investigate how separate topic-specific (i.e. subject, task, etc.) auxiliary memories can be incorporated to store these factorised features, which we believe will further improve the efficiency of knowledge retrieval. (ii) Furthermore, an investigation regarding dynamic auxiliary memory architectures such as graph-structured memory architectures is worthy of consideration. 

\section{Acknowledgement}
The research presented in this paper was supported by the Australian Research Council (ARC) Discovery grant DP200101942.

\bibliographystyle{IEEEtran}
\bibliography{ref}





\end{document}